\documentclass[review]{elsarticle}

\usepackage{lineno,hyperref}
\usepackage{amsmath}
\usepackage{color}

\usepackage{algorithm}
\usepackage{algorithmic}
\usepackage{tabularx}
\usepackage{threeparttable}
\usepackage{epsfig}
\usepackage{amssymb}
\usepackage{pifont}
\usepackage{comment}

\usepackage{multirow}
\usepackage{booktabs}
\usepackage{graphicx}
\usepackage{float}

\newcommand{\xmark}{\ding{55}}%

\journal{Journal of \LaTeX\ Templates}

\begin{document}

\begin{frontmatter}

\title{Self-supervised Feature-Gate Coupling for Dynamic Network Pruning}

\author[MainAddress]{Mengnan Shi\fnref{myfootnote}}
\author[MainAddress]{Chang Liu\fnref{myfootnote}\corref{mycorrespondingauthor}}
\ead{liuchang2022@tsinghua.edu.cn}
\author[SecondaryAddress]{Jianbin Jiao}
\author[SecondaryAddress]{Qixiang Ye}

\cortext[mycorrespondingauthor]{Corresponding author}
\fntext[myfootnote]{M. Shi and C. Liu contribute equally to this work.}

\address[MainAddress]{Tsinghua University, Beijing, 100084, China. E-mail: \{shimn2022,liuchang2022\}@tsinghua.edu.cn}
\address[SecondaryAddress]{University of Chinese Academy of Sciences, Beijing, 100049, China. E-mail: \{jiaojb,qxye\}@ucas.ac.cn}

\begin{abstract}
Gating modules have been widely explored in dynamic network pruning (DNP) to reduce the run-time computational cost of deep neural networks while keeping the features representative.
Despite the substantial progress, existing methods remain ignoring the consistency between feature and gate distributions, which may lead to distortion of gated features. 
In this paper, we propose a feature-gate coupling (FGC) approach aiming to align distributions of features and gates. 
FGC is a plug-and-play module that consists of two steps carried out in an iterative self-supervised manner. In the first step, FGC utilizes the $k$-Nearest Neighbor algorithm in the feature space to explore instance neighborhood relationships, which are treated as self-supervisory signals. 
In the second step, FGC exploits contrastive self-supervised learning (CSL) to regularize gating modules, leading to the alignment of instance neighborhood relationships within the feature and gate spaces. 
Experimental results validate that the proposed FGC method improves the baseline approach with significant margins, outperforming state-of-the-art methods with a better accuracy-computation trade-off. 
Code is publicly available at \href{https://github.com/smn2010/FGC-PR}{\color{magenta}github.com/smn2010/FGC-PR}.
\end{abstract}


\end{frontmatter}
%\linenumbers

%%%%%%%%%%%%%%%%%%% Introduction  %%%%%%%%%%%%%%%%%%%%
\section{Introduction}
%
%
% 
%%%%%%%%%%%%%%%%%%%%%%%%%%%%%%%%%%%%%%%%%%%%%
% 
Convolutional neural networks (CNNs) are becoming deeper to achieve higher performance, bringing overburdened computational costs. Network pruning~\cite{DBLP:conf/cvpr/HeLWHY19, DBLP:conf/cvpr/WuNKRDGF18}, which removes the network parameters of less contribution, has been widely exploited to derive compact network models for resource-constrained scenarios. It aims to reduce the computational cost as much as possible while requiring the pruned network to preserve the representation capacity of the original network for the slightest accuracy drop. 

Existing network pruning methods can be roughly categorized into static and dynamic ones.
Static pruning methods \cite{DBLP:conf/cvpr/HeLWHY19,DBLP:conf/eccv/LiGZGT20, DBLP:journals/pr/LuoW20}, known as channel pruning, derive static simplified models by removing feature channels of negligible performance contribution.
%
% 
%%%%%%%%%%%%%%%%%%%%%%%%%%%%%%%%%%%%%%%%%%%%%%%%%%%%%%
% 
Dynamic pruning methods \cite{DBLP:conf/cvpr/WuNKRDGF18,DBLP:conf/iclr/GaoZDMX19,DBLP:conf/nips/HuaZSZS19} derive input-dependent sub-networks to reduce run-time computational cost. Many dynamic channel pruning methods \cite{DBLP:conf/iclr/GaoZDMX19, ehteshami2020} utilize affiliated gating modules to generate channel-wise binary masks, known as gates, which indicate the removal or preservation of channels. The gating modules explore instance-wise redundancy according to the feature variance of different inputs, $i.e.$, channels recognizing specific features are either turned on or off for distinct input instances.

\begin{figure*}[!t]
    \centering 
    \includegraphics[width=\textwidth]{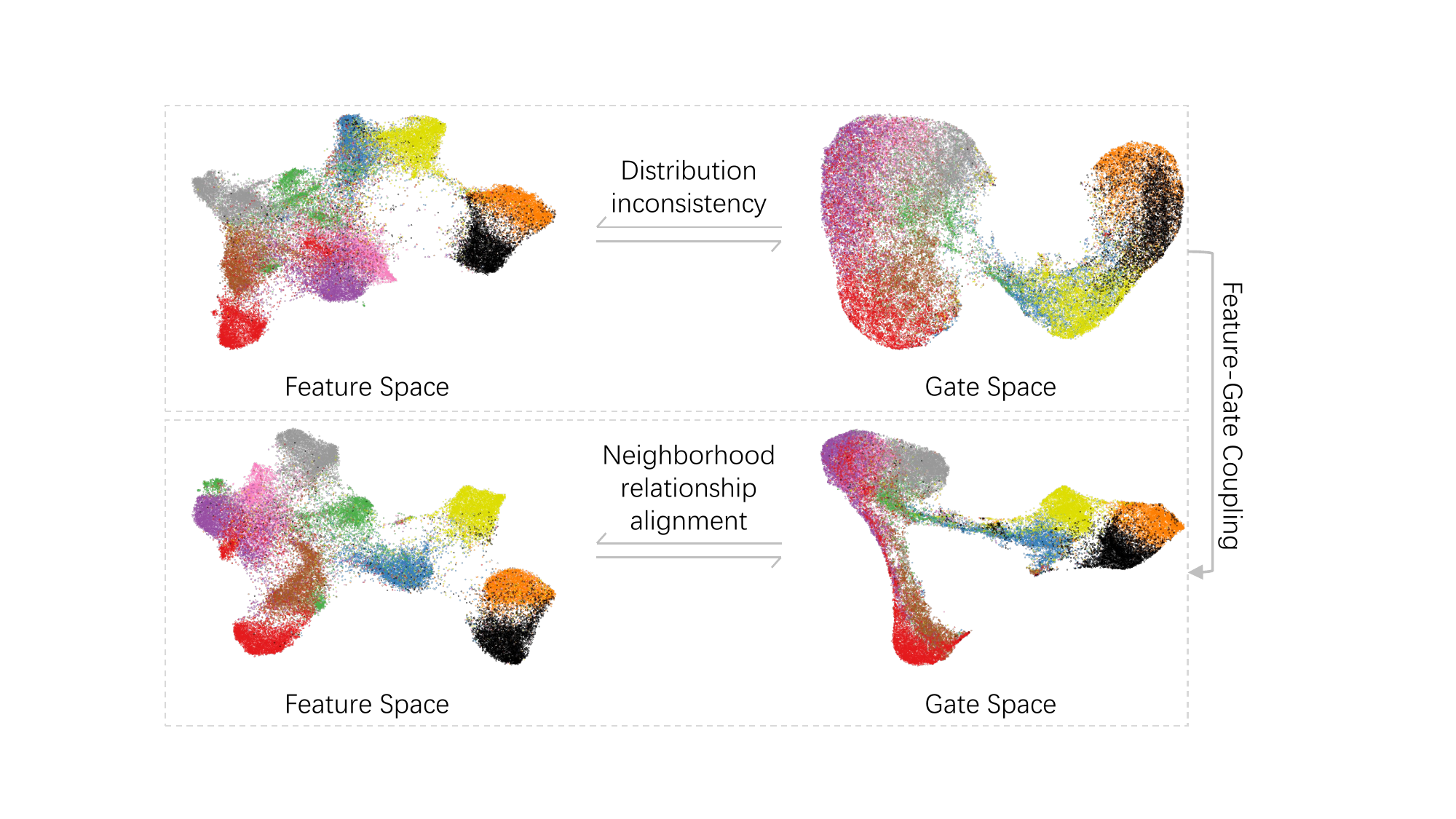}
    \vspace*{-10mm}
    \caption{Up: inconsistent distributions of features and gates without feature-gate coupling (FGC). Down: aligned distributions of features and gates with FGC, \textcolor{black}{where instances with similar semantics would also have similar gates and vice versa. The alignment indicates that FGC allocates consistent sub-networks with specific representation capacities for semantically similar instances, which reduces the representation redundancy and thereby improves the pruning effects.}}
    \label{fig.motivation} 
\end{figure*}

%%%%%%%%%%%%%%% However %%%%%%%%%%%%%%%%%%%%%%%%%%%%%%%%%%%%%
% 
%%%%%%%%%%%%%%%%%%%%%%%%%%%%%%%%%%%%%%%%%%%%%%%%%%%%%%%%%%%%%
Despite substantial progresses, existing methods typically ignore the consistency between feature and gate distributions, $e.g.$, instances with similar features but dissimilar gates, Fig.~\ref{fig.motivation}(upper). 
As gated features are produced by channel-wise multiplication of feature and gate vectors, the distribution inconsistency leads to distortion in the gated feature space. 
The distortion may introduce noise instances to similar instance clusters or disperse them apart, which deteriorates the representation capability of the pruned networks.

%%%%%%%%%%%%%%%%%%%%%%%%%%%%%%%%%%%%%%%%%%
%%%%%%%%%%%%% fig.motivation %%%%%%%%%%%%%
%%%%%%%%%%%%%%%%%%%%%%%%%%%%%%%%%%%%%%%%%%
%%%%%%%%%%%%%%%%%%%%%%%%%%%%%%%%%%%%%%%%%%

%%%%%%%%%%%%%%%%%%%%% In this paper %%%%%%%%%%%%%%%%%%%%
% 
In this paper, we propose a feature-gate coupling (FGC) method to regularize the consistency between distributions of features and gates, Fig.~\ref{fig.motivation}(lower).
The regularization is achieved by aligning neighborhood relationships of instances across feature and gate spaces.
Since ground-truth annotations could not fit the variation of neighborhood relationships in feature spaces across different semantic levels, we propose to fulfill the feature-gate distribution alignment in a self-supervised manner.

Specifically, FGC consists of two steps, which are conducted iteratively. In the first step, FGC utilizes the $k$-Nearest Neighbor ($k$NN) algorithm to explore instance neighborhood relationships in the feature space, where the nearest neighbors of each instance are identified by evaluating feature similarities. The explored neighborhood relationships are then used as self-supervisory signals in the second step, where FGC utilizes contrastive self-supervised learning (CSL)~\cite{DBLP:conf/cvpr/WuXYL18} to regularize gating modules. The nearest neighbors of each instance are identified as positive instances while others as negative instances when defining the contrastive loss function. 
The contrastive loss is optimized in the gate space to pull positive/similar instances together while pushing negative/dissimilar instances away from each other, which guarantees consistent instance neighborhood relationships across the feature and gate spaces.
With gradient back-propagation, parameters in the neural networks are updated as well as feature and gate distributions. FGC then goes back to the first step and conducts the method iteratively until convergence. 
As been designed to perform in the feature and gate spaces, FGC is architecture agnostic and can be applied to various dynamic channel pruning methods in a plug-and-play fashion. 
The contributions of this study are summarized as follows:
\begin{itemize}
%\item[1)]We propose the feature-gate coupling (FGC) method, providing a systematic way to align the distributions of features and corresponding gates of dynamic channel pruning methods in a self-supervised manner.

\item[1)]\textcolor{black}{We propose a plug-and-play feature-gate coupling (FGC) method to alleviate the distribution inconsistency between features and corresponding gates of dynamic channel pruning methods in a self-supervised manner.}

\item[2)]\textcolor{black}{We implement FGC by iteratively performing neighborhood relationship exploration and feature-gate alignment, which aligns instance neighborhood relationships via a designed contrastive loss.}
%\item[2)]We propose a simple-yet-effect approach to implement FGC by iteratively performing neighborhood relationship exploration and feature-gate alignment, reducing the distortion of gated features to a maximum extent.

\item[3)]We conduct experiments on public datasets and network architectures, justifying the efficacy of FGC with significant DNP performance gains.
\end{itemize}

%%%%%%%%%%%%%%%%%%% Related Works %%%%%%%%%%%%%%%%%%%%

\section{Related Works}\label{sec.related_works}

%%%%%%%%%%%%%%%%%%%%%%%%%%%%%%%%%%%%%%%%%%%%%%%%%%%%%%
% dynamic network pruning:
%%%%%%%%%%%%%%%%%%%%%%%%%%%%%%%%%%%%%%%%%%%%%%%%%%%%%%

Network pruning methods can be coarsely classified into static and dynamic ones. \textcolor{black}{Static pruning methods prune neural networks by permanently removing redundant components, such as weights~\cite{DBLP:conf/eccv/LiGZGT20, PR_prune_4, ISTA}, channels~\cite{DBLP:conf/cvpr/HeLWHY19, DBLP:conf/cvpr/LiG0GT20, PR_prune_2, PR_prune_3} or connections~\cite{DBLP:conf/eccv/NingZLLWY20}. They aim to reduce shared redundancy for all instances and derive static compact networks.}
Dynamic pruning methods reduce network redundancy related to each input instance, which usually outperform static methods and have been the research focus of the community. 

\textbf{Dynamic network pruning (DNP)}~\cite{DBLP:conf/cvpr/WuNKRDGF18,DBLP:conf/iclr/GaoZDMX19,DBLP:conf/nips/HuaZSZS19} in recent years has exploited the instance-wise redundancy to accelerate deep networks. The instance-wise redundancy could be roughly categorized into spatial redundancy, complexity redundancy, and parallel representation redundancy. 

Spatial redundancy commonly exists in feature maps where objects may lay in limited regions of the images, while the large background regions contribute marginally to the prediction result. 
Recent methods~\cite{DBLP:conf/nips/HuaZSZS19,DBLP:conf/cvpr/DongHYY17} deal with spatial redundancy to reduce repetitive convolutional computation, or reduce spatial-wise computation in tasks where the spatial redundancy issue becomes more severe due to superfluous backgrounds, such as object detection~\cite{DBLP:conf/eccv/XieZZHL20}, \textit{etc.}

The complexity redundancy has recently deserved attention. As CNNs are getting deeper for recognizing ``harder'' instances, an intuitive way to reduce computational cost is executing fewer layers for ``easier'' instances. Networks could stop at shallow layers~\cite{DBLP:conf/iclr/HuangCLWMW18} once classifiers are capable of recognizing input instances with high confidence. However, the continuous layer removal could be a sub-optimal scheme, especially for ResNet~\cite{7780459} where residual blocks could be safely skipped~\cite{DBLP:conf/cvpr/WuNKRDGF18, DBLP:conf/eccv/VeitB18} without affecting the inference procedure.

Representation redundancy exists in the network because parallel components, $e.g.$, branches and channels, are trained to recognize specific patterns. 
\textcolor{black}{To reduce such redundancy, previous methods~\cite{DBLP:conf/iccv/AhnKO19, PR_dynamic_1} derive dynamic networks using policy networks or neural architecture search methods.}
%
% To reduce such redundancy, previous methods~\cite{DBLP:conf/iccv/AhnKO19} divide channels into several groups and select them with policy networks. 
%
\textcolor{black}{However, they introduce additional overhead and require special training procedures, $i.e.$, reinforcement learning, or neural architecture search.}
%
% However, policy networks introduce additional overhead and require special training procedures, i.e., reinforcement learning. Channel group is not a fine-grained component, where redundancy may be left behind. 
%
% Therefore, 
\textcolor{black}{Recently,} researchers explore pruning channels with affiliated gating modules~\cite{DBLP:conf/iclr/GaoZDMX19, ehteshami2020}, which are small networks with negligible computational overhead. The affiliated gating modules perform the channel-wise selection by gates, reducing redundancy and improving compactness.
\textcolor{black}{Instead of training by only classification loss, \cite{SSMP} defines a self-supervised task to train gating modules, the target masks of which are generated according to corresponding features.}
%%%%%%%%%%%%%%%%%%%%%%%%%%%%%%%%%%%%%%%%%%%%%%%%%%%%%%%%%%
%% 
%%%%%%%%%%%%%%%%%%%%%%%%%%%%%%%%%%%%%%%%%%%%%%%%%%%%%%%%%%
Despite the progress in the reviewed channel pruning methods, the correspondence between gates and features remains ignored, which is the research focus of this study.

%%%%%%%%%%%%%%%%%%%%%%%%%%%%%%%%%%%%%%%%%%%%%%%%%%%%%%%%%%
% Contrastive loss: 
%%%%%%%%%%%%%%%%%%%%%%%%%%%%%%%%%%%%%%%%%%%%%%%%%%%%%%%%%%
\textbf{Contrastive self-supervised learning (CSL)}~\cite{DBLP:conf/cvpr/HadsellCL06} is once used to keep consistent embeddings across spaces, $i.e.$, when a set of high dimensional data points are mapped onto a low dimensional manifold, the ``similar'' points in the original space are mapped to nearby points on the manifold. It achieves this goal by moving similar pairs together and separating dissimilar pairs in the low dimensional manifold, where ``similar'' or ``dissimilar'' points are deemed according to prior knowledge in the high dimensional space.

%%%%%%%%%%%%%%%%%%%%%%%%%%%%%%%%%%%%%%%%%%%%%%%%%%%%%%
%%%%%%%%%%%%%%% fig.2 fgc-pipeline       %%%%%%%%%%%%%
%%%%%%%%%%%%%%%%%%%%%%%%%%%%%%%%%%%%%%%%%%%%%%%%%%%%%%
\begin{figure*}[!t]
    \centering 
    \includegraphics[width=\linewidth]{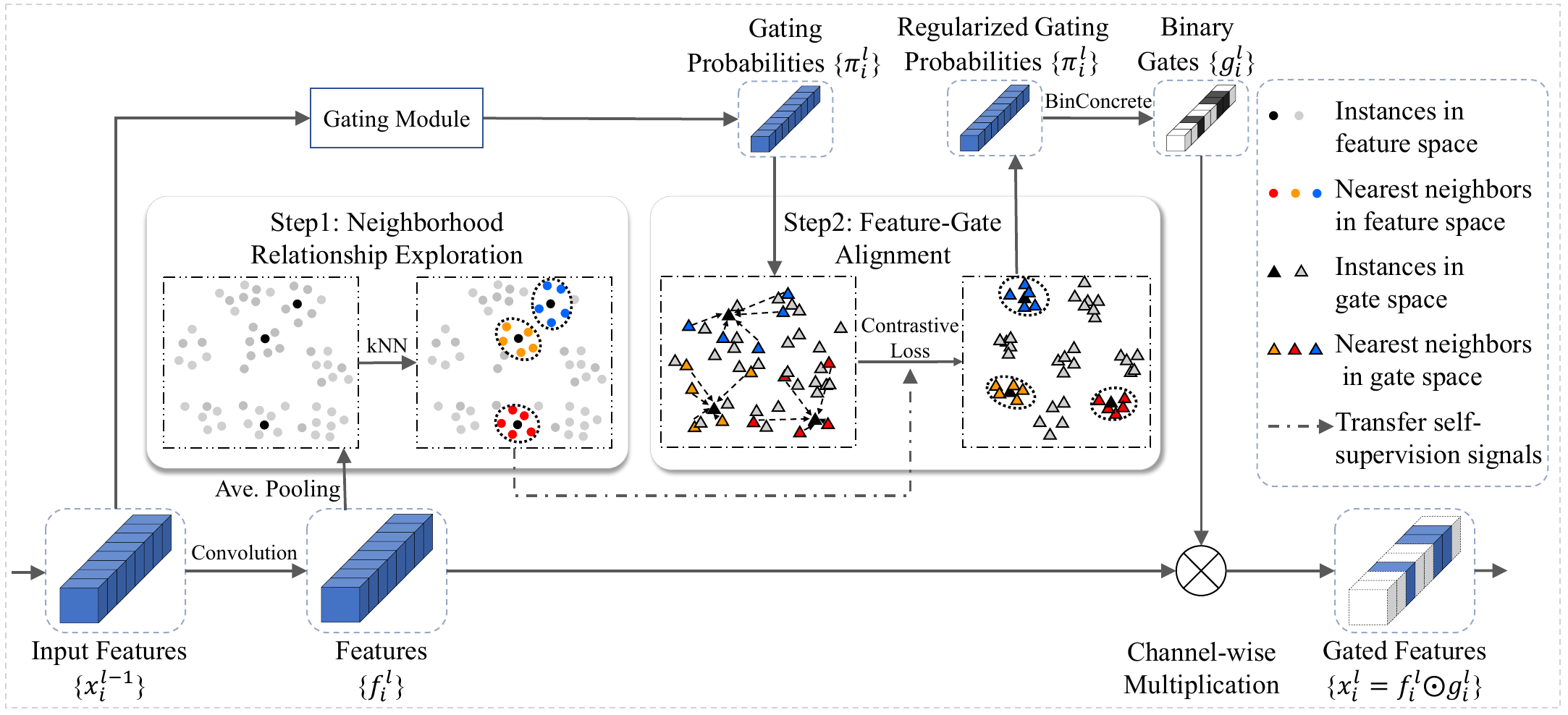} 
    \vspace*{-10mm}
    \caption{Flowchart of the proposed feature-gate coupling (FGC) method for DNP. FGC consists of two iterative steps: neighborhood relationship exploration for instance-wise neighborhood relationship modeling and feature-gate alignment for gating module regularization.}
%}
    \label{fig.pipeline} 
\end{figure*}
%%%%%%%%%%%%%%%%%%%%%%%%%%%%%%%%%%%%%%%%%%%%%%%%%%%%%%

%%%%%%%%%%%%%%%%%%%%%%%%%%%%%%%%%%%%%%%%%%%%%%%%%%%%%%
%% contrastive loss 
%%%%%%%%%%%%%%%%%%%%%%%%%%%%%%%%%%%%%%%%%%%%%%%%%%%%%%
%
CSL has been widely explored in unsupervised and self-supervised learning~\cite{DBLP:conf/cvpr/WuXYL18, DBLP:conf/icml/HuangDGZ19, DBLP:conf/cvpr/He0WXG20} to learn consistent representation. The key point is to identify similar/positive or dissimilar/negative instances. 
An intuitive method is taking each instance itself as positive~\cite{DBLP:conf/cvpr/WuXYL18}. 
However, its relationship with nearby instances is ignored, which is solved by exploring the neighborhood~\cite{DBLP:conf/icml/HuangDGZ19} or local aggregation~\cite{DBLP:conf/iccv/ZhuangZY19}. 
Unsupervised clustering methods~\cite{DBLP:conf/icpr/0004LZW0Y20, DBLP:conf/nips/CaronMMGBJ20} further enlarge the positive sets by treating instances in the same cluster as positives. It is also noticed that positive instances could be gained from multiple views~\cite{DBLP:conf/eccv/TianKI20} of a scene. Besides, recent methods~\cite{DBLP:conf/cvpr/He0WXG20} generate positive instances by siamese encoders, which process data-augmentation of original instances. The assumption is that the instance should stay close in the representation space after whatever augmentation is applied.

%%%%%%%%%%%%%%%%%%%%%%%%%%%%%%%%%%%%%%%%%%
% contrastive loss.
%%%%%%%%%%%%%%%%%%%%%%%%%%%%%%%%%%%%%%%%%%
According to previous works, the contrastive loss is an ideal tool to align distributions or learn consistent representation. Therefore, we develop a CSL method in our work, $i.e.$, the positives are deemed by neighborhood relationships, and the loss function is defined as a variant of InfoNCE~\cite{DBLP:journals/corr/abs-1807-03748}. 

%%%%%%%%%%%%%%%%%%%  Methodology  %%%%%%%%%%%%%%%%%%%%
\section{Methodology} 
%The objective of feature-gate coupling (FGC) is to align distributions of both features and gates to
Inspired by InstDisc~\cite{DBLP:conf/cvpr/WuXYL18}, that visually similar instances are typically close in the feature space, we delve into keeping the distribution alignment between features and gates to avoid distortion after pruning.
To this end, we explore instance neighborhood relationships within the feature space for self-supervisory signals and propose the contrastive loss to regularize gating modules using the generated signals, Fig.~\ref{fig.pipeline}. The two iterative steps can be summarized as follows:
\begin{itemize}
\item \textit{Neighborhood relationship exploration}: We present a $k$NN-based method to explore instance neighborhood relationships in the feature space. Specifically, we evaluate cosine similarities between feature vectors of instance pairs. The similarities are then leveraged to identify the closest instances in the feature space, denoted as nearest neighbors. Finally, indexes of those neighbors are delivered to the next step. 
\item \textit{Feature-gate alignment}: We present a CSL-based method to regularize gating modules by defining positive instances with indexes of neighbors from the previous step. The contrastive loss is then minimized to pull the positives together and push negatives away in the gate space, \textcolor{black}{which achieves feature-gate distribution alignment by improving the consistency between semantic relationships of instances and corresponding gate similarities.}
\end{itemize}

The two steps are conducted iteratively until the optimal dynamic model is obtained. 
\textcolor{black}{With the potential to be applied to other DNP methods, our FGC is inserted into the popular gating module~\cite{ehteshami2020} as an instance.}
%Note that FGC could be generalized to any form of gating module, although it is only tested on a popular gating module~\cite{ehteshami2020} in this study.

%%%%%%%%%%%%%%%%%%%%%%%%%%%%%%%%%%%%%%%%%%%%%%%%%
\subsection{Preliminaries}
\label{sec.pre}
We introduce the basic formulation of gated convolutional layers. For an instance in the dataset $\{x_{i}, y_{i}\}_{n=1}^{N}$, its feature in the $l$-th layer of CNNs is denoted as:
\begin{equation}
f_{i}^{l} = F(x_{i}^{l-1}) = \mathrm{ReLU}(w^{l} \ast x_{i}^{l-1}).
\end{equation}
where $x_{i}^{l-1} \in \mathbb{R}^{C^{l-1}\times W^{l-1}\times H^{l-1}}$ and $f_{i}^{l} \in \mathbb{R}^{C^{l}\times W^{l}\times H^{l}}$ are respectively the input and output of the $l$-th convolutional layer. $C^{l}$ is the channel number. $W^{l}$ and $H^{l}$ represent the width and height of the feature map. $w^{l}$ denotes the convolutional weight matrix. $\ast$ is the convolution operator, and $\rm{ReLU}$ is the activation function. Batch normalization is applied after each convolution layer. %\textcolor{red}{H?} 

Dynamic channel pruning methods temporarily remove channels to reduce related computation in both preceding and succeeding layers, where a binary gate vector, generated from the preceding feature $x_{i}^{l-1}$, is used to turn on or off corresponding channels. The gated feature in the $l$-th layer is denoted as:
\begin{equation}
    x_{i}^{l} = g_{i}^{l}\odot f_{i}^{l},
\label{eq.gated}
\end{equation}
where $g_{i}^{l}$ denotes the gate vector and $\odot$ denotes the channel-wise multiplication.

The gate vectors come from various gating modules with uniform functionality. Without loss of generality, we use the form of gating module~\cite{ehteshami2020}, which is defined as:
\begin{equation}
    g_{i}^{l} = G(x_{i}^{l-1})= B(M(p(x_{i}^{l-1}))).
\label{eq.bas}
\end{equation}
The gating module first goes with the global average pooling layer $p$, which is used to generate the spatial descriptor of the input feature $x_{i}^{l-1}$. A lightweight network \textcolor{black}{$M$} with two fully connected layers is then applied to generate the gating probability $\pi_{i}^{l}=Linear(p(x_{i}^{l-1}))$. Finally, \textcolor{black}{the \textit{BinConcrete}~\cite{DBLP:conf/iclr/JangGP17} activation function $B$} is used to sample from $\pi_{i}^{l}$ to get the binary mask $g_{i}^{l}$.

\subsection{Implementation}
\label{sec.implementation}
\textbf{Neighborhood Relationship Exploration.}
%%%%%%%%%%%%%%%%%%%%%%%%%%%%%%%%%%%%%%%%%%%%%%%%%
%%%%%%%%%%%%%%%%%%%%%%%%%%%%%%%%%%%%%%%%%%%%%%%%%
Instance neighborhood relationships vary in feature spaces with different semantic levels, $e.g$,  in a low-level feature space, instances with similar colors or texture may be closer, while in a high-level feature space, instances from the same class may gather together. Thus, human annotations could not provide adaptive supervision of instance neighborhood relationships across different network stages. We propose to utilize the $k$NN algorithm to adaptively explore instance neighborhood relationships in each feature space and extract self-supervision accordingly for feature-gate distribution alignment.

%%%%%%%%%%%%%%%%%%%%%%%%%%%%%%%%%%%%%%%%%%%%%%%%%
% kNN
%%%%%%%%%%%%%%%%%%%%%%%%%%%%%%%%%%%%%%%%%%%%%%%%%
Note that the exploration could be adopted in arbitrary layers, $e.g.$, the $l$-th layer of CNNs. In practice, in the $l$-th layer, the feature $f_{i}^{l}$ of the $i$-th instance is first fed to a global average pooling layer, as $\hat{f}_{i}^{l}=p(f_{i}^{l})$. The dimension of pooled feature reduces from $C^{l}\times W^{l}\times H^{l}$ to $C^{l}\times 1\times1$, which is more efficient for processing. We then compute its similarities with other instances in the training dataset as:
\begin{equation}
    \mathcal{S}^{l}[i, j] = \hat{f}_{i}^{lT} \cdot \hat{f}_{j}^{l},
\label{eq.similarity}
\end{equation}
where $\mathcal{S}^{l} \in \mathbb R^{N \times N}$ denotes the similarity matrix and $\mathcal{S}^{l}[i, j]$ denotes similarity between the $i$-th instance and the $j$-th instance. $N$ denotes the number of total training instances. To identify nearest neighbors of $x_{i}$, we sort the $i$-th row of $\mathcal{S}_{l}$ and return the subscripts of the $k$ biggest elements, as
\begin{equation}
    \mathcal{N}_{i}^{l} = \mathop{topk}(\mathcal{S}^{l}[i, :]).
\label{eq.topk}
\end{equation}
The set of subscripts $\mathcal{N}_{i}^{l}$, $i.e.$, indexes of neighbor instances in the training dataset, are treated as self-supervisory signals to regularizing gating modules.

To compute the similarity matrix $\mathcal{S}^{l}$ in Eq.~(\ref{eq.similarity}), $\hat{f}_{j}^{l}$ for all instances are required. Instead of exhaustively computing these pooled features at each iteration, we maintain a feature memory bank $\mathcal{V}_{f}^{l}\in \mathbb{R}^{N \times D}$ to store them~\cite{DBLP:conf/cvpr/WuXYL18}. $D$ is the dimension of the pooled feature, $i.e.$, the number of channels $C^{l}$. Now for the $i$-th pooled feature $\hat{f}_{i}^{l}$, we compute its similarities with vectors in $\mathcal{V}_{f}^{l}$. After that, the memory bank is updated in a momentum way, as
\begin{equation}
    \mathcal{V}_{f}^{l}[i] \leftarrow m * \mathcal{V}_{f}^{l}[i] + (1-m) * \hat{f}_{i}^{l},
\label{eq.momentum_update}
\end{equation}
where the momentum coefficient is empirically set to 0.5. All vectors in the memory bank are initialized as unit random vectors.

\textbf{Feature-Gate Alignment.} 
We identify the nearest neighbor set $\{x_{j}, j \in \mathcal{N}_{i}^{l}\}$ of each instance $x_{i}$ in the feature space. Thereby, instances in each neighbor set are treated as positives to be pulled closer while others as negatives to be pushed away in the gate space. To this end, we utilize the contrastive loss in a self-supervised manner.

Specifically, by defining nearest neighbors of an instance as its positives, the probability of an input instance $x_{j}$ being recognized as nearest neighbors of $x_{i}$ in the gate space, as
\begin{equation}
    p(j|\pi_{i}^{l}) = \frac {\exp(\pi_{j}^{lT} \cdot \pi_{i}^{l}/\tau)}{\sum_{k=1}^{N}(\exp(\pi_{k}^{lT}\cdot \pi_{i}^{l})/\tau)},
\label{eq.prob}
\end{equation}
where $\pi_{i}^{l}$ and $\pi_{j}^{l}$ are gating probabilities, the gating module outputs of $x_{i}$ and $x_{j}$ respectively. $\tau$ is a temperature hyper-parameter and set as 0.07 according to \cite{DBLP:conf/cvpr/WuXYL18}. Then the loss function is derived after summing up $p(j|\pi_{i}^{l})$ of all $k$ neighbors, as
\begin{equation}
    \mathcal{L}_{g}^{l} = -\sum_{j \in \mathcal{N}_{i}^{l}}{\log(p(j|\pi_{i}^{l}))}.
\label{eq.contrast}
\end{equation}

%%%%%%%%%%%%%%%%%%%%%%%%%%%%%%%%%%%%%%%%%%%%%
% contrastive loss
%%%%%%%%%%%%%%%%%%%%%%%%%%%%%%%%%%%%%%%%%%%%%
Since nearest neighbors are defined as positives and others as negatives, Eq.~(\ref{eq.contrast}) is minimized to draw previous nearest neighbors close, $i.e.$, reproducing the instance neighborhood relationships within the gate space.
%
%%%%%%%%%%%%%%%%%%%%%%%%%%%%%%%%%%%%%%%%%%%%%
% gate memory bank
%%%%%%%%%%%%%%%%%%%%%%%%%%%%%%%%%%%%%%%%%%%%%
We maintain a gate memory bank $\mathcal {V}_{g}^{l} \in \mathbb{R}^{N \times D}$ to store gating probabilities. For each $\pi_{i}^{l}$, its positives are taken from $\mathcal {V}_{g}^{l}$ according to $\mathcal{N}_{i}^{l}$. After computing Eq.~(\ref{eq.contrast}), $\mathcal {V}_{g}^{l}$ is updated in a similar momentum way as updating the feature memory bank. All vectors in $\mathcal {V}_{g}^{l}$ are initialized as unit random vectors.

%%%%%%%%%%%%%%%%%%%%%%%%%%%%%%%%%%%%%%%%
%%%%         algorithm           %%%%%%%
%%%%%%%%%%%%%%%%%%%%%%%%%%%%%%%%%%%%%%%%
\begin{algorithm}[ht] 
    \caption{Feature-Gate Coupling.}
    \label{algorithm1} 
\begin{algorithmic}[1]
\REQUIRE Training instances $x_{i}$, labels $y_{i}$ of an imagery dataset;
\ENSURE Dynamic network with parameters $\theta$;

\STATE Initialize feature and gate memory banks $\mathcal{V}^{l}_{f}$ and $\mathcal{V}^{l}_{g}$ for layers in $\Omega$;
\\

\FOR{layer $l$ = 1 {\bf to} $L$}
    \STATE Generate feature $f_{i}$ and gating probability $\pi_{i}$ of instance $x_{i}$;
    \IF{$l \in \Omega$}
        \STATE Generate pooled feature $\hat{f}_{i}$ of instance $x_{i}$;
    
        \COMMENT {\textbf{Step 1. Neighborhood Relationship Exploration}}
        \STATE Calculate the similarities of $\hat{f}^{l}_{i}$ with $\mathcal{V}^{l}_{f}[j], (j\in(1, N), j\neq i$), (Eq.~(\ref{eq.similarity}))
        \STATE Get indexes of nearest neighbors $\mathcal {N}^{l}_{i}$, (Eq.~(\ref{eq.topk}));
        \STATE Update the feature memory bank $\mathcal{V}^{l}_{f}$, (Eq.~(\ref{eq.momentum_update}));
        \\
        \COMMENT{\textbf{Step 2. Feature-Gate Alignment}}
        \STATE Draw positive instances w.r.t. $\mathcal {N}^{l}_{i}$ from the gate memory bank $\mathcal{V}^{l}_{g}$;
        \STATE Calculate the contrastive loss in the gate space $\mathcal{L}^{l}_{g}$, (Eq.~(\ref{eq.contrast}));
        \STATE Update the gate memory bank $\mathcal{V}^{l}_{g}$;
    \ENDIF
\STATE Calculate $\mathcal{L}^{l}_{0}$;
\ENDFOR
\STATE Calculate the overall objective loss, (Eq.~(\ref{eq.loss}));
\STATE Updating parameters $\theta$ by gradient back-propagation.

\end{algorithmic}
\end{algorithm}

%%%%%%%%%%%%%%%%%%%%%%%%%%%%%%%%%%%%%%%%
% Overall Objective Loss
%%%%%%%%%%%%%%%%%%%%%%%%%%%%%%%%%%%%%%%%
\textbf{Overall Objective Loss.}
Besides the proposed contrastive loss, we utilize the cross-entropy loss ($\mathcal{L}_{ce}$) for the classification task and the $\mathcal L_{0}$ loss to introduce sparsity. 
$\mathcal{L}_{0}^{l} = \lVert \pi_{i}^{l} \rVert_{0}$ is used in each gated layer, where $\lVert \cdot \rVert_{0}$ is $\mathcal L_{0}$-norm.

The overall loss function of our method can be formulated as:
\begin{equation}
    loss = \mathcal{L}_{ce} + \eta \cdot \sum_{l \in \Omega} \mathcal{L}_{g}^{l} + \rho \cdot \sum_{l} \mathcal{L}_{0}^{l},
\label{eq.loss}
\end{equation}
where $\mathcal{L}_{g}^{l}$ denotes the contrastive loss adopted in the $l$-th layer and $\eta$ the corresponding coefficient, $\Omega$ denotes the layers $\mathcal{L}_{g}^{l}$ applied in. The coefficient $\rho$ controls the sparsity of pruned models, $i.e.$, a bigger $\rho$ leads to a sparser model with lower accuracy, while a lower $\rho$ derives less sparsity and higher accuracy.

The learning procedure of FGC is summarized as Algorithm \ref{algorithm1}.

%%%%%%%%%%%%%%%%%%%%%%%%%%%%%%%%%%%%%%%%
%%%%%%%%%%%%%%%%%%%%%%%%%%%%%%%%%%%%%%%%
\subsection{Theoretical Analysis}
\label{sec.mi}
We analyze the plausibility of designed contrastive loss (Eq.~(\ref{eq.contrast})) from the perspective of mutual information maximization. According to~\cite{DBLP:journals/corr/abs-1807-03748}, the InfoNCE loss is used to maximize the mutual information between learned representation and input signals. It is assumed that minimizing the contrastive loss leads to the improved mutual information between features and corresponding gates.

We define the mutual information between gates and features as:
\begin{equation}
    I(f^{l}, g^{l}) = \sum_{f^{l}, g^{l}} p(f^{l}, g^{l})\log\frac{p(f^{l}|g^{l})}{p(f^{l})},
    \label{eq.mi}
\end{equation}
\textcolor{black}{where the density ratio based on $f_{j}^{l}$ and $g_{i}^{l}$ is proportional to a positive real score as $q(f_{j}^{l}, g_{i}^{l}) \propto p(f_{j}^{l}|g_{i}^{l})/p(f_{j}^{l})$~\cite{DBLP:journals/corr/abs-1807-03748}, which is $q(f_{j}^{l}, g_{i}^{l}) = \exp(\pi_{j}^{lT} \cdot \pi_{i}^{l}/\tau)$ in our case. $\pi_{i}^{l}$ is the gating probabilities for $g_{i}^{l}$ and $\pi_{j}^{l}$ belongs to its positives. The InfoNCE loss item could be rewritten as:}

%The mutual information between $f_{j}^{l}$ and $g_{i}^{l}$ can be modeled by a density ratio $q(f_{j}^{l}, g_{i}^{l}) \propto p(f_{j}^{l}|g_{i}^{l})/p(f_{j}^{l})$, and in our case $q(f_{j}^{l}, g_{i}^{l}) = \exp(\pi_{j}^{lT} \cdot \pi_{i}^{l}/\tau)$, as Eq.~(\ref{eq.prob}) shows. $\pi_{i}^{l}$ is the gating probabilities for $g_{i}^{l}$ and $\pi_{j}^{l}$ belongs to its positives. The InfoNCE loss function is then rewritten as:
\begin{equation}
    \mathcal{L}_{g}^{l} = - \mathop{\mathbb{E}}_{X}[\log\frac{ q(f_{j}^{l},g_{i}^{l})}{\sum_{f_{k}^{l}\in f^{l}} q(f_{k}^{l}, g_{i}^{l})}].
\end{equation}
\textcolor{black}{By taking the density ratio into the equation, we could derive the relationship between InfoNCE loss and mutual information, where the superscript $l$ is omitted for brevity,}
\begin{equation}
\begin{split}
    \mathcal{L}_{NCE} 
    &= - \mathop{\mathbb{E}}_{X}\log[\frac{ q(f_{j},g_{i})}{\sum_{f_{k}} q(f_{k}, g_{i})}]
    \\
    &= - \mathop{\mathbb{E}}_{X}\log[\frac{\frac{p(f_j|g_i)}{p(f_j)}}{\frac{p(f_j|g_i)}{p(f_j)}+\sum_{f_k\in X_{neg}}\frac{p(f_k|g_i)}{p(f_k)}}]
    \\
    &= \mathop{\mathbb{E}}_{X}\log[1 + \frac{p(f_j)}{p(f_j|g_i)}\sum_{f_k\in X_{neg}}\frac{p(f_k|g_i)}{p(f_k)}]
    \\
    &\approx \mathop{\mathbb{E}}_{X} \log[1 + \frac{p(f_j)}{p(f_j|g_i)}(N-1)\mathop{\mathbb{E}}_{f_k}\frac{p(f_k|g_i)}{p(f_k)}]
    \\
    &= \mathop{\mathbb{E}}_{X} \log[1 + \frac{p(f_j)}{p(f_j|g_i)}(N-1)]
    \\
    &= \mathop{\mathbb{E}}_{X} \log[\frac{p(f_j|g_i)-p(f_j)}{p(f_j|g_i)} + \frac{p(f_j)}{p(f_j|g_i)}N]
    \\
    &\geq \mathop{\mathbb{E}}_{X} \log[ \frac{p(f_j)}{p(f_j|g_i)}N]
    \\
    &= -I(f_j,g_i) + \log(N),
\end{split}
\end{equation}
\textcolor{black}{where $N$ is the number of training samples, and $X_{neg}$ denotes the set containing negative instances. The inequality holds because $f_j$ and $g_i$ is not independent and $p(f_j|g_i) \geq p(f_j)$.}

\textcolor{black}{As $I(f_{j}, g_{i}) \geq \mathop{\log}(N)-\mathcal{L}_{NCE}$, we could conclude that minimizing the InfoNCE loss item would maximize a lower bound on mutual information, justifying the efficacy of proposed contrastive loss $\mathcal{L}_{g}^{l} = \sum_{j \in \mathcal{N}_{i}^{l}}\mathcal{L}_{NCE}^{l}$.} Accordingly, the improvement of mutual information between features and corresponding gates facilities distribution alignment. (Please refer to Fig.~\ref{fig.nmi}.)
%where fljf_{j}^{l} denotes nearest neighbors in the feature space.
%
%The mutual information between fljf_{j}^{l} and glig_{i}^{l} is evaluated as:
% \begin{equation}
%     I(f_{j}^{l}, g_{i}^{l}) \geq \mathop{\log}(N)-\mathcal{L}_{g}^{l}.
% \end{equation}
% It indicates that minimizing the InfoNCE loss Llg\mathcal{L}_{g}^{l} maximizes a lower bound of mutual information. Accordingly, the improvement of mutual information between features and corresponding gates facilities distribution alignment. (Please refer to Fig.~?????????\ref{fig.nmi}.)

%%%%%%%%%%%%%%%%%%%%%%%%%%%%%%%%%%%%%%%%%%%%%%%%%%%%%%%%%%%%%%%%%%%%%%%%%%%%%%%%%%%%%%%%%%%%%%%%%%%%%%%%%%%%%
\section{Experiments}
In this section, the experimental settings for dynamic channel pruning are first described. Ablation studies are then conducted to validate the effectiveness of FGC. The alignment of instance neighborhood relationships by the FGC is also quantitatively and qualitatively analyzed. We show the performance of learned dynamic networks on image classification benchmark datasets and compare them with state-of-the-art pruning methods. Finally, We conduct object detection and semantic segmentation experiments to show the generation ability of FGC.

\subsection{Experimental Settings}
\textbf{Datasets.} Experiments on image classification are conducted on benchmarks including CIFAR10/100~\cite{cifar10} and ImageNet~\cite{Deng2009ImageNetAL}. CIFAR10 and CIFAR100 consist of 10 and 100 categories, respectively. Both datasets contain 60K colored images, 50K and 10K of which are used for training and testing. ImageNet is an important large-scale dataset composed of 1000 object categories, which has 1.28 million images for training and 50K images for validation.

{We conduct object detection on PASCAL VOC 2007~\cite{Everingham2009Pascal}, the trainval and test sets of which are used for training and testing, respectively. Instance segmentation is conducted on Cityscapes~\cite{Cordts2016Cityscapes}. We use the train set for training and the validation set for testing.
}

\textbf{Implementation Details.} For CIFAR10 and CIFAR100, training images are applied with standard data augmentation schemes~\cite{ehteshami2020}. We use the ResNet of various depths, $i.e.$, ResNet--\{20, 32, 56\}, to fully demonstrate the effectiveness of our method. On CIFAR10, models are trained for 400 epochs with a mini-batch of 256. A Nesterov’s accelerated gradient descent optimizer is used, where the momentum is set to 0.9 and the weight decay is set to 5e-4. It is noticed that no weight decay is applied to parameters in gating modules. As for the learning rate, we adopt a multiple-step scheduler, where the initial value is set to 0.1 and decays by 0.1 at milestone epochs [200, 275, 350]. 
\textcolor{black}{We also evalutate FGC on WideResNet~\cite{WideResNet} and MobileNetV2~\cite{MobileNetV2} (a type of lightweight network). Specifically, models are trained for 200 epochs with a mini-batch of 256. The initial learning rate is 0.1 and decays by 0.2 at epochs [60, 120, 160].}
On CIFAR100, models are trained for 200 epochs with a mini-batch of 128. The same optimization settings as CIFAR10 are used. The initial learning rate is set to 0.1, then decays by 0.2 at epochs [60, 120, 160].
For ImageNet, the data augmentation scheme is adopted from~\cite{7780459}. We only use ResNet-18 to validate the effectiveness on the large-scale dataset due to time-consuming training procedures. We train models for 130 epochs with a mini-batch of 256. A similar optimizer is used except for the weight decay of 1e-4. The learning rate is initialized to 0.1 and decays by 0.1 at epochs [40, 70, 100].
FGC is employed in the last few residual blocks of ResNets, $i.e.$, the last two blocks of ResNet-\{20,32\}, the last four blocks of ResNet-56, and the last block of ResNet-18. Hyper-parameters $\eta$ and $k$ are set to 0.003 and 200 by default, respectively. We set the l0 loss coefficient $\rho$ as 0.4 to introduce considerable sparsity into gates.
{For PASCAL VOC 2007, we use the detector and the data augmentation described in Faster R-CNN~\cite{Ren2015FasterRCNN}. The Faster R-CNN model consists of the ResNet backbone and Feature Pyramid Network~\cite{Lin2017FPN} (FPN). Specifically, We replace the backbone with the gated ResNet-\{34, 50\} pre-trained on ImageNet, leaving other components unchanged. We fine-tune the detector under the 1x schedule where the initial learning rate is 0.001, the momentum is 0.9, and the weight decay is 1e-4. No weight decay is applied in gating modules. FGC is employed in the last stage of the backbone. During the fine-tuning process, hyper-parameters $\eta$ and $k$ are set to 0.0001 and 1, respectively. }

%%%%%%%%%%%%%%%%%%%%%%%%%%%%%%%%%%%%%%%%%%%%%%%%%%%%
%%%%%%%%%%%%%%%   tab.k  & fig.k  %%%%%%%%%%%%%%%%%%
%%%%%%%%%%%%%%%%%%%%%%%%%%%%%%%%%%%%%%%%%%%%%%%%%%%%
\begin{table}[t]
\footnotesize
\centering
\caption{Test errors and pruning ratios of the pruned ResNet-20 w.r.t the number of nearest neighbors $k$. ``Error'' is the classification error, where ``$\downarrow$'' denotes the lower the better. ``Pruning'' is the pruning ratio of computation, where ``$\uparrow$'' denotes the higher the better. (Ditto for other tables.)}
\label{tab.K}
    \begin{tabular}{@{}ccccccccc@{}}
    \toprule
    $k$                    & 5    & 20   & 100  & 200            & 512   & 1024  & 2048  & 4096  \\ \midrule
    Error (\%) $\downarrow$ & 8.47 & 8.01 & 8.00 & \textbf{7.91} & 8.16 & 8.04 & 8.09 & 8.39 \\
    Pruning (\%) $\uparrow$ & 50.1 & 54.1 & 53.4 & \textbf{55.1} & 53.3 & 54.9 & 53.0 & 53.8 \\ \bottomrule
    \end{tabular}%
\end{table}

\begin{figure}[t]
\centering 
\includegraphics[width=\textwidth]{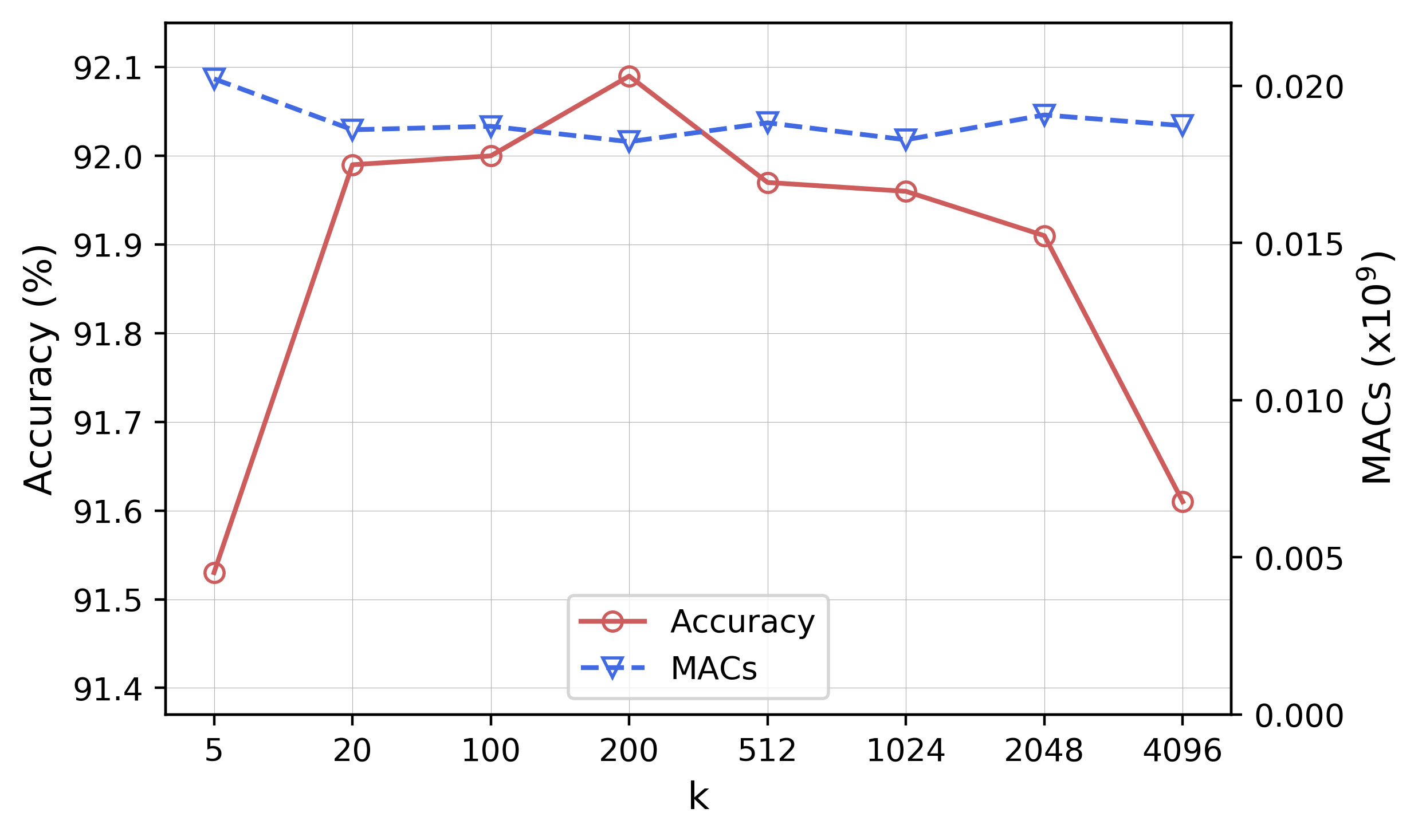}
\vspace*{-10mm}
\caption{Test accuracies and computation of the pruned ResNet-20 w.r.t the number of nearest neighbors $k$.}
\label{fig.K} 
\end{figure}
%%%%%%%%%%%%%%%%%%%%%%%%%%%%%%%%%%%%%%%%%%%%%%%%%%%%

% 
{For Cityscapes, we follow the data augmentation and the method as PSPNet~\cite{Zhao2017PSPnet}. PSPNet with a pre-trained ResNet-50 backbone is trained with a mini-batch of 6 for 300 epochs. The optimizer is an SGD optimizer with the poly learning rate schedule, where the initial learning rate is 0.02 and the decay exponent is 0.9. The weight decay and momentum are 5e-4 and 0.9, respectively. The weight decay of gating modules is 2e-6. In practice, We employ FGC in the last stage of the backbone. We set $\eta$ and $k$ to 0.0001 and 4.}

%
%%%%%%%%%%%%%%%%%%%%%%%%%%%%%%%%%%%%%%%%%%%%%%%%%%%%
%%%%%%%%%%%%%  tab.eta  & fig.eta  %%%%%%%%%%%%%%%%%
%%%%%%%%%%%%%%%%%%%%%%%%%%%%%%%%%%%%%%%%%%%%%%%%%%%%
% \begin{table}[t]
%     \footnotesize
%     \centering
%     \setlength{\belowcaptionskip}{2pt}
%     \caption{Test errors and pruning ratios of the pruned ResNet-20 w.r.t coefficient η\eta.}
%     \label{tab.eta}
%         \begin{tabular}{@{}cccccccccc@{}}
%         \toprule
%         η\eta                  & 0    & 5e-4 & 0.001 & 0.002  & 0.003  &\textcolor{black}{0.004} & 0.005    & 0.01 & 0.02 \\ \midrule
%         Error (\%) ↓\downarrow & 8.59 & 8.47 & 8.22  & 8.12          & \textbf{7.91}  & \textcolor{black}{8.04} & 7.94     & 8.01 & 8.21 \\
%         Pruning (\%) ↑\uparrow & 54.0 & 54.8 & 53.2  & \textbf{55.3} & 55.1           & \textcolor{black}{54.5} & 53.1     & 53.0 & 52.4 \\ \bottomrule
%         \end{tabular}%
% \end{table}
\begin{table}
    \footnotesize
    \centering
    \caption{Test errors and pruning ratios of the pruned ResNet-20 w.r.t coefficient $\eta$.}
    \begin{tabular}{cccccccc}
        \toprule
        $\eta$ ($\times$1e-3) & 0 & 0.5 & 1 & 1.5 & 2 & 2.5 & 3 \\
        \midrule
        Error (\%) $\downarrow$ & 8.59 & 8.47 & 8.22 & 8.14 & 8.12 & 8.11 & 7.91 \\
        Pruning (\%) $\uparrow$ & 5.40 & 54.8 & 53.2 & 53.9 & 55.3 & 54.8 & 55.1 \\
        \hline
        $\eta$ ($\times$1e-3) & 3.5 & 4 & 4.5 & 5 &10 &20\\
        \midrule
        Error (\%) $\downarrow$ & 8.07 & 8.04 & 8.18 & 7.94 & 8.01 & 8.21 \\
        Pruning (\%) $\uparrow$ & 54.9 & 54.5 & 53.7 & 53.1 & 53.0 & 52.4 \\
        \bottomrule
    \end{tabular}
    \label{tab.eta}
\end{table}

\begin{figure*}[t]
\centering 
\includegraphics[width=\textwidth]{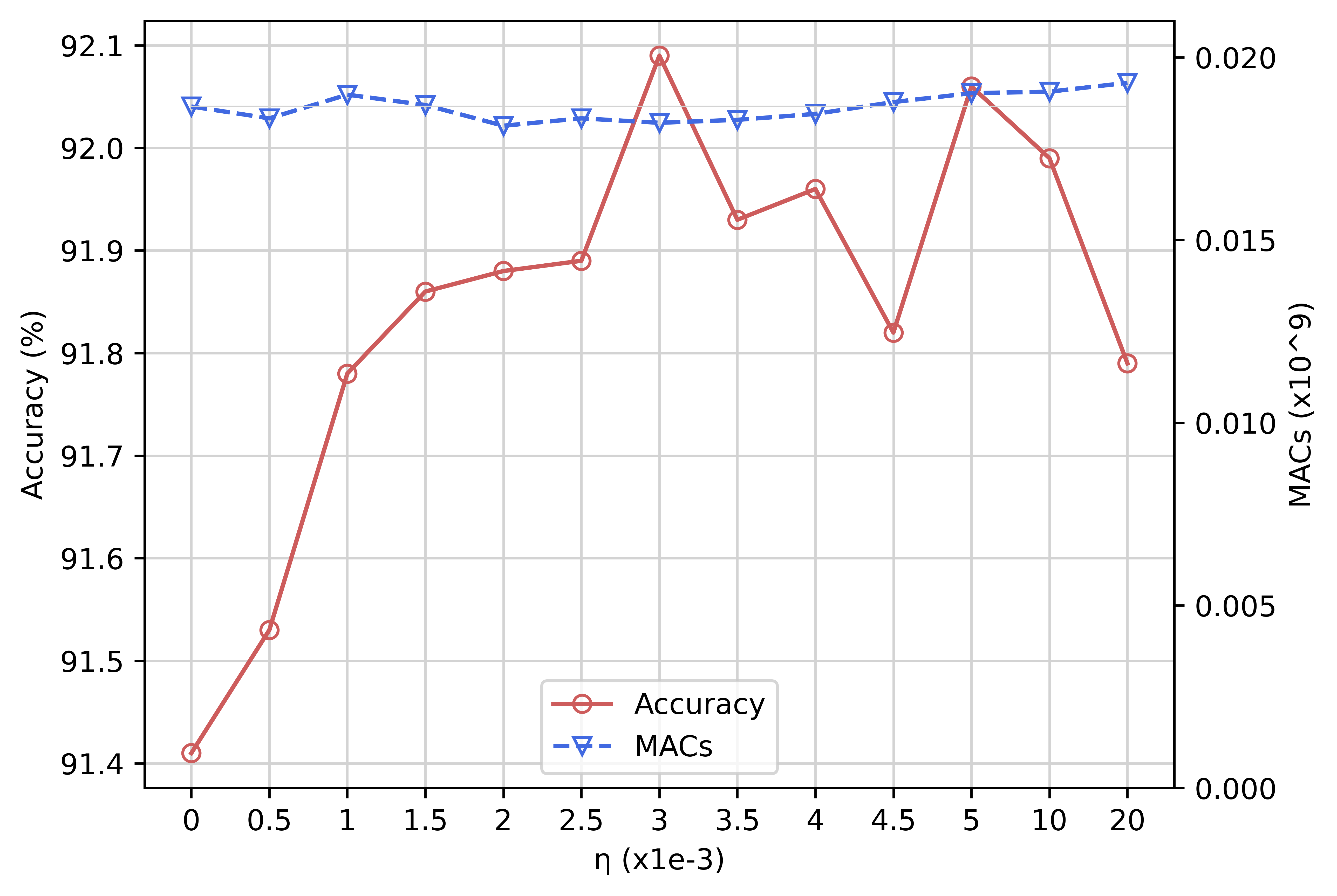}
\vspace*{-10mm}
\caption{Test accuracies and computation of the pruned ResNet-20 w.r.t coefficient $\eta$.}
\label{fig.eta} 
\end{figure*}
%%%%%%%%%%%%%%%%%%%%%%%%%%%%%%%%%%%%%%%%%%%%%%%%%%%%

\subsection{Ablation studies}
We conduct ablation studies on CIFAR10 to validate the effectiveness of our method by exploring hyper-parameters, residual blocks, neighborhood relationships, and accuracy-computation trade-offs.

\textbf{Hyper-parameters of Contrastive Loss.}
In Table \ref{tab.K} and Fig.~\ref{fig.K}, we show the performance of pruned models given different $k$, $i.e.$, the number of nearest neighbors. For a fair comparison, the computation of each model is tuned to be roughly the same so that it can be evaluated by accuracy. As shown by Fig.~\ref{fig.K}, FGC is robust about $k$ in a wide range, $i.e.$, the accuracy reaches the highest value around $k=$200, and keeps stable when $k$ falls into the range (20, 1024). 

In Table \ref{tab.eta} and Fig. \ref{fig.eta}, we show the performance of pruned models under different $\eta$, $i.e.$, the loss coefficient. Similarly, the computation of each model is also tuned to be almost the same. The contrastive loss brings a significant accuracy improvement compared with models without FGC. With the increase of $\eta$, the accuracy improves from 91.4\% to 92.09\%. The best performance is achieved when $\eta=0.003$. While accuracy slightly drops when $\eta$ is larger than 0.01, it remains better than that at $\eta=0$. 

\textbf{Residual Blocks.}
In Table~\ref{tab.layer}, we exploit whether FGC is more effective when employed in multiple residual blocks. The first row of Table~\ref{tab.layer} denotes models trained without FGC. After gradually removing shallow residual blocks from adopting FGC, experimental results show that FGC in shallow stages brings a limited performance gain despite more memory banks, $e.g.$, total $18$ memory banks used when employing FGC in all blocks. However, FGC in the last stage, especially from the $8$-$th$ to the $9$-$th$ blocks, contributes better to the final performance, $i.e.$, the lowest error (7.91\%) with considerable computation reduction (55.1\%). It is concluded that the residual blocks in deep stages benefit more from FGC, probably due to more discriminative features and more conditional gates. Therefore, taking both training efficiency and performance into consideration, we employ FGC in the deep stages of ResNets.

%%%%%%%%%%%%%%%%%%%%%%%%%%%%%%%%%%%%%%%%%%%%%%%%%%%%
%%%%%%%%          tab.layers              %%%%%%%%%%
%%%%%%%%%%%%%%%%%%%%%%%%%%%%%%%%%%%%%%%%%%%%%%%%%%%%
\begin{table}[t]
    \footnotesize
    \centering
    \setlength{\belowcaptionskip}{2pt}
    \caption{Test errors and pruning ratios of the pruned ResNet-20 when employing FGC in different blocks. \textcolor{black}{For ResNet-20, which consists of three stages, each of them comprising three basic residual blocks, we use indexes of ``Stages'' and ``Blocks'' to locate the application of FGC.}}
    \label{tab.layer}
        \begin{tabular}{@{}ccccc@{}}
        \toprule
        Stages                    & Blocks                   & Memory Banks   & Error (\%) $\downarrow$ & Pruning (\%) $\uparrow$    \\ \midrule
        -                         & -                         & -     & 8.49               & 54.3    \\    \hline
        $1^{st}$ to $3^{rd}$      & $1^{st}$ to $9^{th}$ & 18    & 8.08                    & 53.4    \\    \hline
        $2^{nd}$ to $3^{rd}$      & $4^{th}$ to $9^{th}$ & 12    & 8.36                    & \textbf{56.0}  \\ \hline
        \multirow{3}{*}{$3^{rd}$} & $7^{th}$ to $9^{th}$ & 6     & 8.21                    & \textbf{56.0}  \\
                                  & $8^{th}$ to $9^{th}$ & 4     & \textbf{7.91}           & 55.1    \\
                                  & Only $9^{th}$        & 2     & 8.14                    & 54.5     \\ \bottomrule
        \end{tabular}%
\end{table}
%%%%%%%%%%%%%%%%%%%%%%%%%%%%%%%%%%%%%%%%%%%%%%%%%%%%

\textbf{Neighborhood Relationship Sources.} In Table~\ref{tab.source}, we validate the necessity of exploring instance neighborhood relationships in feature spaces by comparing the counterparts. Specifically, we randomly select $k$ instances with the same label as neighbors for each input instance, denoted as ``Label''; we also utilize $k$NN in the gate space to generate nearest neighbors, denoted as ``Gate''. Moreover, since FGC is applied in multiple layers separately, an alternative adopts neighborhood relationships from the same layer, $e.g.$, the last layer. We define this case as ``Feature'' with ``Shared'' instance neighborhood relationships.

Results in Table~\ref{tab.source} demonstrate that our method, $i.e.$, ``Feature'' with independent neighbors, prevails over all the counterparts, which fail to explore proper instance neighborhood relationships for alignment. For example, labels are too coarse or discrete to describe neighborhood relationships. 
Exploring neighbors in the gate space enhances its misalignment with the feature space. Therefore the performance, either error or pruning ratio (8.50\%, 51.6\%), is even worse than the model without FGC. Moreover, ``Shared'' neighbors from the same feature space could not deal with variants of intermediate features, thereby limiting the improvement.

We conclude that the proposed FGC consistently outperforms the compared methods, which attributes to the alignment of instance neighborhood relationships between feature and gate spaces in multiple layers.

%%%%%%%%%%%%%%%%%%%%%%%%%%%%%%%%%%%%%%%%%%%%%%%%%%%%
%%%%%%%%%          tab.neighbors.      %%%%%%%%%%%%%
%%%%%%%%%%%%%%%%%%%%%%%%%%%%%%%%%%%%%%%%%%%%%%%%%%%%
\begin{table}
    \footnotesize
    \centering
    \setlength{\belowcaptionskip}{2pt}
    \caption{Comparisons of neighbor sources with ResNet-20 for instance neighborhood relationship exploration.}
    \label{tab.source}
        \begin{tabular}{@{}cccc@{}}
        \toprule
        Neighbor Source      & Shared         & Error (\%) $\downarrow$ & Pruning (\%) $\uparrow$   \\ \midrule
        -           & -              & 8.49                    & 54.3          \\        \hline
        Label       & \xmark         & 8.16                    & 52.0          \\
        Gate        & \xmark         & 8.50                    & 51.6          \\
        Feature     & \checkmark     & 8.14                    & 51.2          \\        \hline
        Feature     & \xmark         & \textbf{7.91}           & \textbf{55.1} \\        \bottomrule
        \end{tabular}%
\end{table}
%%%%%%%%%%%%%%%%%%%%%%%%%%%%%%%%%%%%%%%%%%%%%%%%%%%%

\begin{figure}[!t]
    \centering 
    \includegraphics[width=\textwidth]{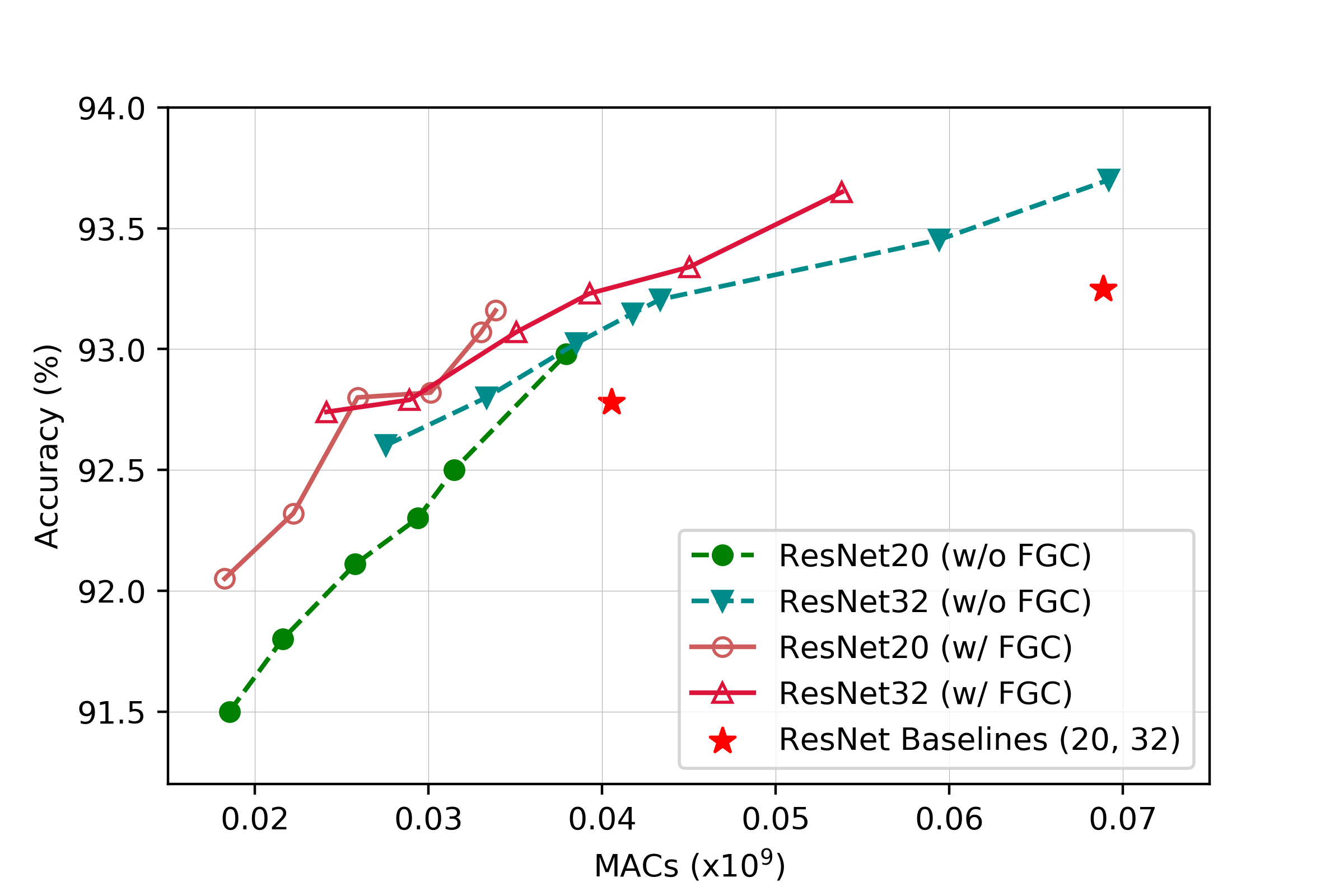}
    \vspace*{-10mm}
    \caption{\textcolor{black}{Computation-accuracy trade-offs of pruned ResNet-\{20, 32\}. The ``w/'' and ``w/o'' refer to ``with'' and ``without'', respectively, indicating whether the FGC method is used or not. (Ditto for other tables or figures.)}}
    \label{fig.rho} 
\end{figure}

\textbf{Computation-Accuracy Trade-offs.} 
In Fig.~\ref{fig.rho}, we show the computation-accuracy trade-offs of FGC and the baseline method by selecting $\mathcal L_{0}$ loss coefficient $\rho$ from values of \{0.01, 0.02, 0.05, 0.1, 0.2, 0.4\}. Generally speaking, the accuracy of pruned models increases as the total computational cost increases. Compared with pruned models without FGC, the pruned ResNet-32 with FGC obtains higher accuracy with lower computational cost, as shown in the top right of Fig.~\ref{fig.rho}. At the left bottom of Fig.~\ref{fig.rho}, the pruned ResNet-20 with FGC achieves ~$0.5\%$ higher accuracy using the same computational cost. The efficacy of FGC is clarified that increasing performances compared with the baseline models can be observed at lower pruning ratios. 

%%%%%%%%%%%%%%%%%%%%%%%%%%%%%%%%%%%%%%%%%%%%%%%%%%%%
%%%%%%%%%%%%%%%%%  fig.trade-off  %%%%%%%%%%%%%%%%%%
%%%%%%%%%%%%%%%%%%%%%%%%%%%%%%%%%%%%%%%%%%%%%%%%%%%%
%%%%%%%%%%%%%%%%%%%%%%%%%%%%%%%%%%%%%%%%%%%%%%%%%%%%

\subsection{Analysis}

\begin{figure*}[!t]
    \centering
    \includegraphics[width=\textwidth]{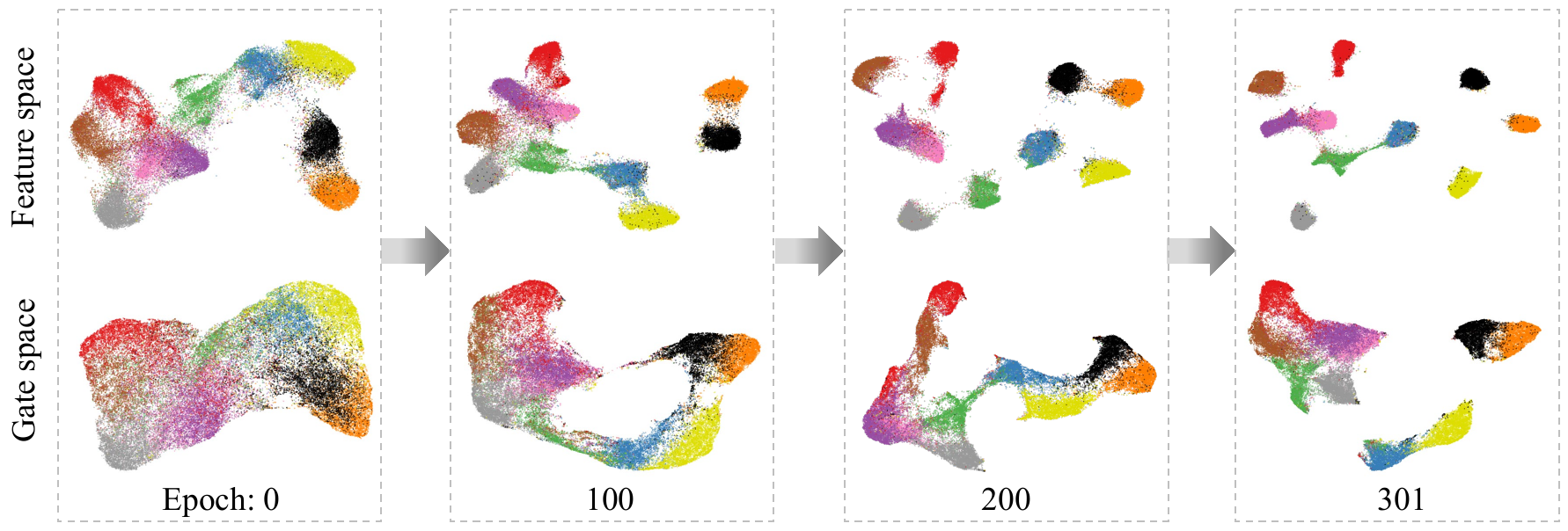}
    \vspace*{-10mm}
    \caption{Embedding visualization of features and gates in ResNet-20 (conv3-3) trained with FGC at different training epochs. Each point denotes an instance in the dataset (CIFAR10). Each color denotes one category. (Best view in color).}
    \label{fig.process} 
\end{figure*}

\textbf{Visualization Analysis.} In Fig.~\ref{fig.process}, we use UMAP~\cite{DBLP:journals/corr/abs-1802-03426} to visualize the correspondence of features and gates during the training procedure. Specifically, we show features and gates from the last residual block (conv3-3) of ResNet-20, where features are trained to be discriminative, $i.e.$, easy to be distinguished in the embedding space. It's seen that features and gates are simultaneously optimized and evolve together during training. In the last epoch, as instance neighborhood relationships have been well reproduced in the gate space, we can observe gates with a similar distribution as features, $e.g.$, instances of the same clusters in the feature space are also assembled in the gate space. Intuitively, the coupling of features and gates promises better predictions of our pruned models with given pruning ratios.

%%%%%%%%%%%%%%%%%%%%%%%%%%%%%%%%%%%%%%%%%%%%%%%%%%%%
%%%%%%%%%%%%%%%%%%%  fig.process  %%%%%%%%%%%%%%%%%%
%%%%%%%%%%%%%%%%%%%%%%%%%%%%%%%%%%%%%%%%%%%%%%%%%%%%
%%%%%%%%%%%%%%%%%%%%%%%%%%%%%%%%%%%%%%%%%%%%%%%%%%%%

%%%%%%%%%%%%%%%%%%%%%%%%%%%%%%%%%%%%%%%%%%%%%%%%%%%%
%%%%%%%%%%%%%     fig.comparison   %%%%%%%%%%%%%%%%%
%%%%%%%%%%%%%%%%%%%%%%%%%%%%%%%%%%%%%%%%%%%%%%%%%%%%
\begin{figure*}[!t]
    \centering 
    \includegraphics[width=\textwidth]{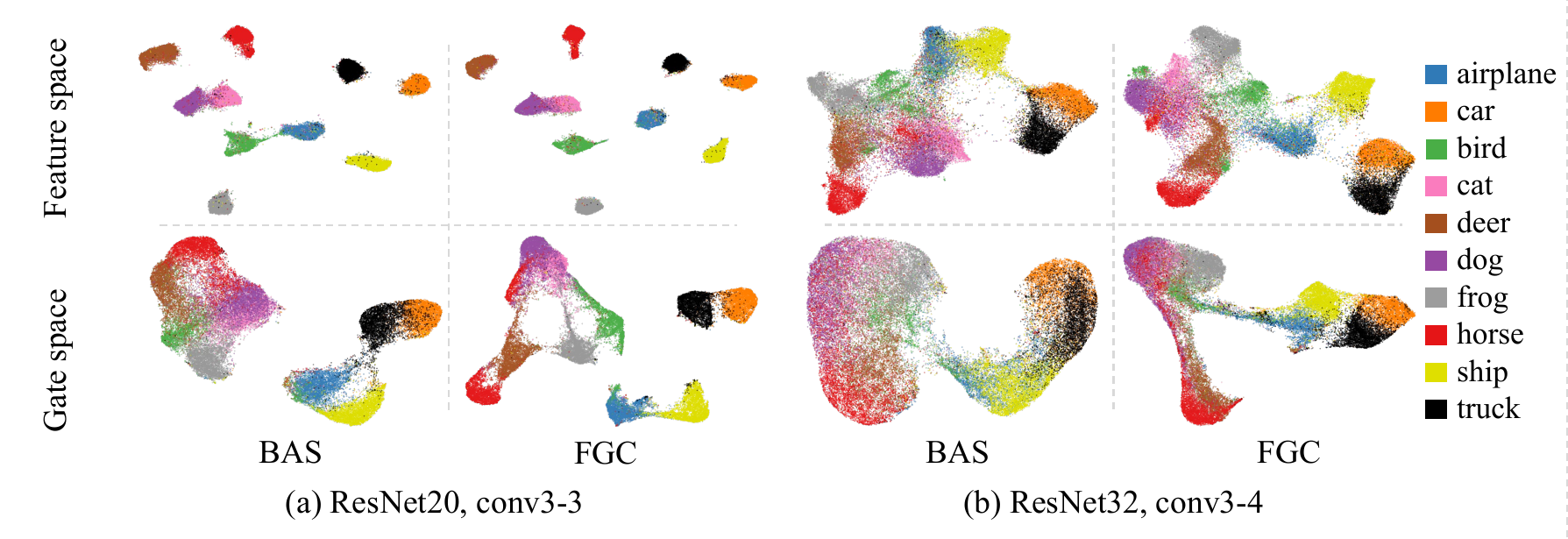}
    \vspace*{-10mm}
    \caption{Embedding visualization of features and gates in pruned models trained with FGC, compared with pruned models in BAS~\cite{ehteshami2020}. (Best view in color).}
    \label{fig.comparison} 
\end{figure*}
%%%%%%%%%%%%%%%%%%%%%%%%%%%%%%%%%%%%%%%%%%%%%%%%%%%%

In Fig.~\ref{fig.comparison}, we compare feature and gate embeddings with pruned models in BAS~\cite{ehteshami2020}. Specifically, they are drawn from the last residual block (conv3-3) of ResNet-20 and the second to last residual block (conv3-4) of ResNet-32. As shown, our method brings pruned models with coupled features and gates, compared with BAS. For example, ``bird'' (green), ``frog'' (grey) and ``deer'' (brown) instances entangle together (left bottom of Fig.~\ref{fig.comparison}a), while adopting FGC (right bottom of Fig.~\ref{fig.comparison}a), they are clearly distinguished as corresponding features. Similar observations are seen in Fig.~\ref{fig.comparison}b, \textcolor{black}{validating the alignment effect of FGC, which facilitates the model to leverage specific inference paths for semantically consistent instances and thereby decrease representation redundancy.}

\begin{figure*}[!t]
\centering 
\includegraphics[width=\textwidth]{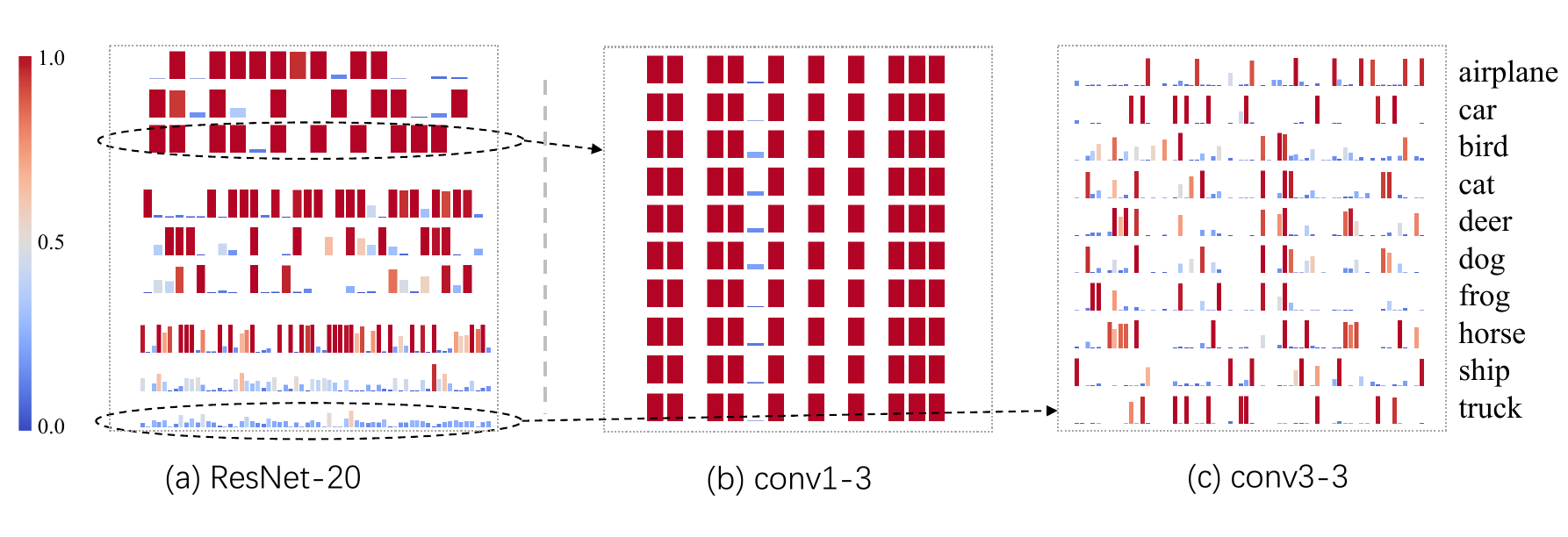}
\vspace*{-10mm}
\caption{Execution frequency of channels in pruned models. The color and height of each bar denote the value of execution frequency. (a) Execution frequency in each layer listed from top to bottom. (b) Execution frequency of each category in conv1-3. (c) Execution frequency of each category in conv3-3.}
\label{fig.patterns} 
\end{figure*}

In Fig.~\ref{fig.patterns}, we visualize the execution frequency of channels in pruned ResNet-20 for each category on CIFAR10. In shallow layers ($e.g.$, conv1-3), gates almost entirely turn on or off for arbitrary categories, indicating why FGC is unnecessary: a common channel compression scheme is enough. However, in deep layers ($e.g.$, conv3-3), dynamic pruning is performed according to the semantics of categories. It's noticed that semantically similar categories, such as ``dog'' and ``cat'', also share similar gate execution patterns. 
We conclude that FGC utilizes and enhances the correlation between instance semantics and pruning patterns to pursue better representation.

%%%%%%%%%%%%%%%%%%%%%%%%%%%%%%%%%%%%%%%%%%%%%%
%%%%%%%%%%%%%%%   fig.pattern   %%%%%%%%%%%%%%
%%%%%%%%%%%%%%%%%%%%%%%%%%%%%%%%%%%%%%%%%%%%%%
%%%%%%%%%%%%%%%%%%%%%%%%%%%%%%%%%%%%%%%%%%%%%%

\begin{figure*}[!t]
    \centering 
    \includegraphics[width=1\linewidth]{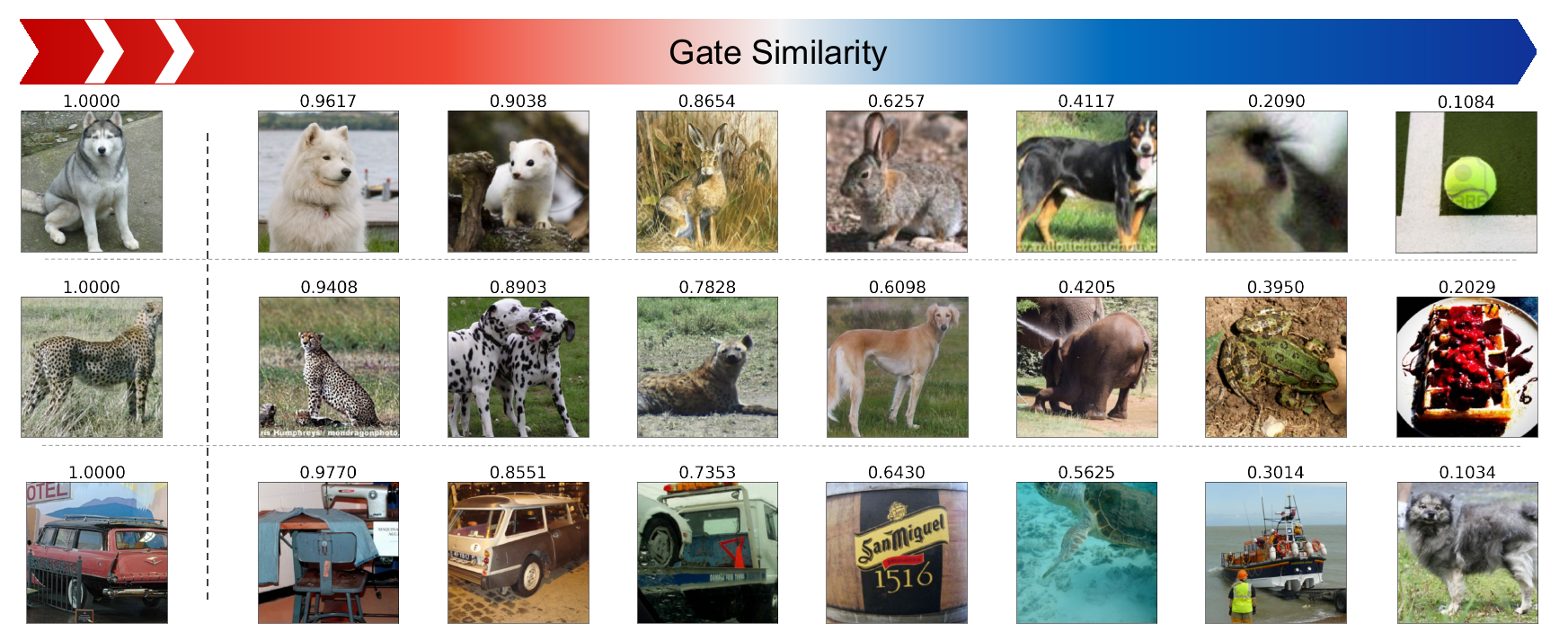}
    \vspace*{-10mm}
    \caption{Sorted images w.r.t gate similarities on ImageNet validation set. In each row, we list example images according to their gate similarities (values above images) with the leftmost one. Gates are drawn from the $8$-$th$ gated block of ResNet-18.}
    \label{fig.examples} 
\end{figure*}

In Fig.~\ref{fig.examples}, we show images in the ordering of gate similarities with the target ones (leftmost). One can see that the semantic similarities and gate similarities of image instances decrease consistently from left to right. 

%%%%%%%%%%%%%%%%%%%%%%%%%%%%%%%%%%%%%%%%%%%%%%%%%%%
%%%%%%%%%%%%%%%% fig. examples %%%%%%%%%%%%%%%%%%%%
%%%%%%%%%%%%%%%%%%%%%%%%%%%%%%%%%%%%%%%%%%%%%%%%%%%
%%%%%%%%%%%%%%%%%%%%%%%%%%%%%%%%%%%%%%%%%%%%%%%%%%%

\begin{figure*}[!t]
    \centering 
    \includegraphics[width=1\linewidth]{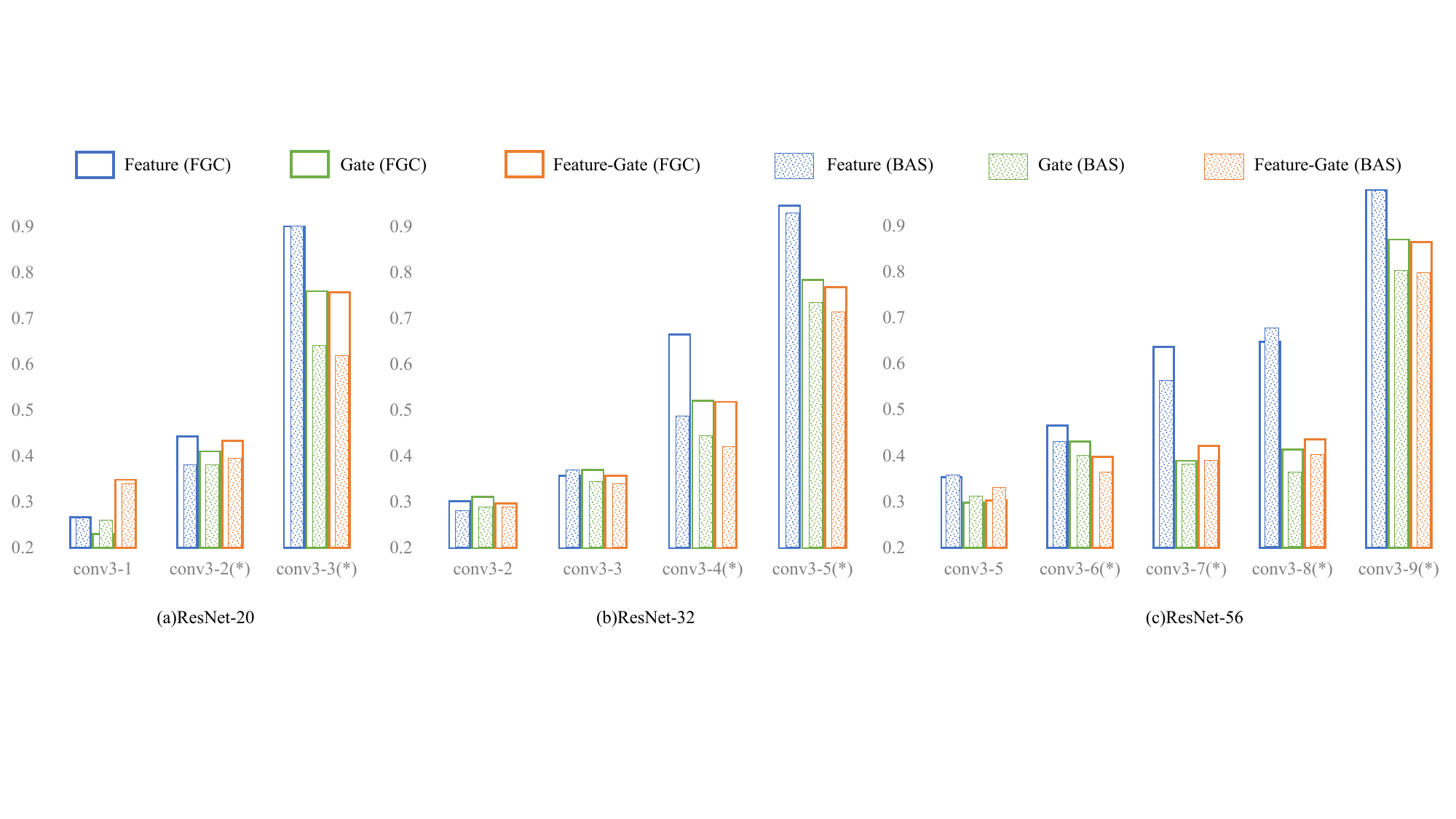}
    \vspace*{-10mm}
    \caption{Normalized mutual information (NMI) in deep residual blocks of ResNet-\{20, 32, 56\}, compared with pruned models in BAS~\cite{ehteshami2020}. ``Feature'' denotes NMI between features and labels; ``Gate'' denotes NMI between gates and labels; ``Feature-Gate'' denotes NMI between features and gates. ``$\ast$'' denotes residual blocks employed with FGC.}
    \label{fig.nmi}
\end{figure*}

\textbf{Mutual Information Analysis.}
%%%%%%%%%%%%%%%%%%%%%%%%%%%%%%%%%%%%%%%%%%%%%%%%%%%
In Fig.~\ref{fig.nmi}, we validate that FGC improves the mutual information between features and gates to facility distribution alignment, as discussed in section~\ref{sec.mi}. For comparison, only partial deep layers are employed with FGC in our pruned models, indicated by ``$\ast$''. Specifically, the normalized mutual information (NMI) between features and gates is roughly unchanged for layers not employed with FGC.
However, we observe the consistent improvement of NMI between features and gates for layers employed with FGC compared with the pruned models in BAS~\cite{ehteshami2020}.

%%%%%%%%%%%%%%%%%%%%%%%%%%%%%%%%%%%%%%%%%%%%%%%%%%%
%%%%%%%%%%%%%%%%%%%  fig.nmi  %%%%%%%%%%%%%%%%%%%%%
%%%%%%%%%%%%%%%%%%%%%%%%%%%%%%%%%%%%%%%%%%%%%%%%%%%
%%%%%%%%%%%%%%%%%%%%%%%%%%%%%%%%%%%%%%%%%%%%%%%%%%%

\subsection{Performance}
\textcolor{black}{
We compare the proposed method with state-of-the-art methods, including static and dynamic methods. For a clear and comprehensive comparison, we classify these methods into different types $w.r.t.$ pruned objects. Specifically, most static methods rely on pruning channels (or pruning filters, which is equivalent), while some prune weights in filter kernels or connections between channels. Dynamic methods are classified into three types, layer pruning, spatial pruning, and channel pruning, as discussed in Sec.~\ref{sec.related_works}. The performance is evaluated by two metrics, classification errors and average pruning ratios of computation. Higher pruning ratios and lower errors indicate better-pruned models.}

%%%%%%%%%%%%%%%%%%%%%%%%%%%%%%%%%%%%%%%%%%%%%%%
%%%%%%%%%%%%%%%   CIFAR10   %%%%%%%%%%%%%%%%%%%
%%%%%%%%%%%%%%%%%%%%%%%%%%%%%%%%%%%%%%%%%%%%%%%
\begin{table}[htp]
%\small
\scriptsize
%\tiny
\centering
\setlength{\belowcaptionskip}{2pt}
\caption{Comparisons of the pruned ResNet on CIFAR-10. \textcolor{black}{Pruning methods are classified into different types w.r.t. pruned objects, i.e., ``\textbf{C}" for channels, ``CN" for connections, ``W" for weights, ``\textbf{L}" for layers, ``\textbf{S}" for spatial. (Ditto for other tables.)}}
\label{tab:cifar10}
\begin{tabular}{@{}cccccc@{}}
\toprule
Model & Method & Dynamic & \textcolor{black}{Type} & Error (\%) $\downarrow$ & Pruning (\%) $\uparrow$ \\ \midrule 
\multirow{13}{*}{\rotatebox[origin=c]{90}{ResNet-20}} & Baseline   & -  & \textcolor{black}{-}  & 7.22          & 0.0           \\
                     & SFP~\cite{DBLP:conf/ijcai/HeKDFY18}   & \xmark  & \textcolor{black}{C}  & 9.17          & 42.2          \\
                     & FPGM~\cite{DBLP:conf/cvpr/HeLWHY19}   & \xmark   & \textcolor{black}{C}  & 9.56         & 54.0          \\
                     & DSA~\cite{DBLP:conf/eccv/NingZLLWY20} & \xmark   & \textcolor{black}{CN}  & 8.62         & 50.3          \\
                     & Hinge~\cite{DBLP:conf/cvpr/LiG0GT20}  & \xmark   & \textcolor{black}{C}  & 8.16         & 45.5          \\
                     & DHP~\cite{DBLP:conf/eccv/LiGZGT20}    & \xmark   & \textcolor{black}{W}  & 8.46          & 51.8          \\
                     & \textcolor{black}{AIG~\cite{DBLP:conf/eccv/VeitB18}}    & \textcolor{black}{\checkmark} & \textcolor{black}{L} & \textcolor{black}{8.49}         & \textcolor{black}{24.7}
                            \\
                     & \textcolor{black}{SACT~\cite{DBLP:conf/eccv/XieZZHL20}}  & \textcolor{black}{\checkmark} & \textcolor{black}{S}  & \textcolor{black}{10.02}       & \textcolor{black}{53.7}
                            \\
                     & \textcolor{black}{DynConv~\cite{Verelst2020DynamicCE}}   & \textcolor{black}{\checkmark} & \textcolor{black}{S}  & \textcolor{black}{8.10}        & \textcolor{black}{50.7}
                            \\
                     & FBS~\cite{DBLP:conf/iclr/GaoZDMX19}   & \checkmark & \textcolor{black}{C} & 9.03        & 53.1          \\
                     & BAS~\cite{ehteshami2020}              & \checkmark & \textcolor{black}{C} & 8.49        & 54.3          \\ \hline
                     & \textbf{FGC (Ours)}               & \checkmark & \textcolor{black}{C} & \textbf{7.91}  & 55.1           \\
                     & \textbf{FGC (Ours)}               & \checkmark & \textcolor{black}{C} & 8.13          & \textbf{57.4}   \\ \midrule \midrule
\multirow{11}{*}{\rotatebox[origin=c]{90}{ResNet-32}}  & Baseline & - & \textcolor{black}{-}   & 6.75          & 0.0           \\
                     & SFP~\cite{DBLP:conf/ijcai/HeKDFY18}   & \xmark & \textcolor{black}{C}  & 7.92          & 41.5          \\
                     & FPGM~\cite{DBLP:conf/cvpr/HeLWHY19}   & \xmark & \textcolor{black}{C}  & 8.07          & 53.2          \\
                     & \textcolor{black}{AIG~\cite{DBLP:conf/eccv/VeitB18}}     & \textcolor{black}{\checkmark} & \textcolor{black}{L}  & \textcolor{black}{8.49}       & \textcolor{black}{24.7}
                          \\
                     & \textcolor{black}{BlockDrop~\cite{DBLP:conf/cvpr/WuNKRDGF18}} & \textcolor{black}{\checkmark} & \textcolor{black}{L} & \textcolor{black}{8.70} & \textcolor{black}{54.0}
                          \\
                     & \textcolor{black}{SACT~\cite{DBLP:conf/eccv/XieZZHL20}}  & \textcolor{black}{\checkmark} & \textcolor{black}{S}  & \textcolor{black}{8.00}       & \textcolor{black}{38.4}
                          \\
                     & \textcolor{black}{DynConv~\cite{Verelst2020DynamicCE}}   & \textcolor{black}{\checkmark} & \textcolor{black}{S}  & \textcolor{black}{7.47}       & \textcolor{black}{51.3}
                          \\                
                     & FBS~\cite{DBLP:conf/iclr/GaoZDMX19}   & \checkmark & \textcolor{black}{C} & 8.02       & 55.7          \\
                     & BAS~\cite{ehteshami2020}              & \checkmark & \textcolor{black}{C} & 7.79       & 64.2          \\ \hline
                     & \textbf{FGC (Ours)}                   & \checkmark & \textcolor{black}{C} & \textbf{7.26} & 65.0          \\
                     & \textbf{FGC (Ours)}                   & \checkmark & \textcolor{black}{C} & 7.43  & \textbf{66.9}       \\ \midrule \midrule
\multirow{13}{*}{\rotatebox[origin=c]{90}{ResNet-56}} & Baseline   & -   & \textcolor{black}{-}   & 5.86          & 0.0           \\
                     & SFP~\cite{DBLP:conf/ijcai/HeKDFY18}    & \xmark  & \textcolor{black}{C}   & 7.74          & 52.6          \\
                     & FPGM~\cite{DBLP:conf/cvpr/HeLWHY19}    & \xmark  & \textcolor{black}{C}   & 6.51          & 52.6          \\
                     & HRank~\cite{DBLP:conf/cvpr/LinJWZZ0020}& \xmark  & \textcolor{black}{C}   & 6.83          & 50.0          \\
                     & DSA~\cite{DBLP:conf/eccv/NingZLLWY20}  & \xmark  & \textcolor{black}{CN}   & 7.09          & 52.2          \\
                     & Hinge~\cite{DBLP:conf/cvpr/LiG0GT20}   & \xmark  & \textcolor{black}{C}   & 6.31          & 50.0          \\
                     & DHP~\cite{DBLP:conf/eccv/LiGZGT20}     & \xmark  & \textcolor{black}{W}    & 6.42          & 50.9          \\
                     & \textcolor{black}{ResRep~\cite{Ding2021ResRepLC}}        & \textcolor{black}{\xmark}  & \textcolor{black}{C}   & \textcolor{black}{6.29}          & \textcolor{black}{52.9}
                             \\
                     & \textcolor{black}{RL-MCTS~\cite{Wang2022ChannelPV}}      & \textcolor{black}{\xmark}  & \textcolor{black}{C}   & \textcolor{black}{6.44}         & \textcolor{black}{55.0}
                             \\
                     & FBS~\cite{DBLP:conf/iclr/GaoZDMX19}    & \checkmark & \textcolor{black}{C} & 6.48         & 53.6          \\
                     & BAS~\cite{ehteshami2020}               & \checkmark & \textcolor{black}{C} & 6.43         & 62.9          \\ \hline
                     & \textbf{FGC (Ours)}      & \checkmark & \textcolor{black}{C} & \textbf{6.09} & 62.2          \\
                     & \textbf{FGC (Ours)}      & \checkmark & \textcolor{black}{C} & 6.41          & \textbf{66.2} \\ 
                     \bottomrule
\end{tabular}%
\end{table}
%%%%%%%%%%%%%%%%%%%%%%%%%%%%%%%%%%%%%%%%%%%%%%%

\textcolor{black}{
\textbf{CIFAR10.} In Table~\ref{tab:cifar10}, comparisons with SOTA methods to prune ResNet-\{20, 32, 56\} are shown. FGC significantly outperforms static pruning methods. For example, our dynamic model for ResNet-56 reduces $66.2\%$ computation, $11.2\%$ more than RL-MCTS~\cite{Wang2022ChannelPV}, given a rather similar accuracy~($6.41\%$ $vs$ $6.44\%$). The reason lies in that FGC further reduces the instance-wise redundancy of input samples.}

\textcolor{black}{
It prevails various types of dynamic pruning techniques. Specifically, FGC significantly outperforms AIG~\cite{DBLP:conf/eccv/VeitB18} and Blockdrop~\cite{DBLP:conf/cvpr/WuNKRDGF18}, typical dynamic layer pruning methods that derive dynamic inference by skipping layers. AIG for ResNet-20 achieves a slightly higher error than FGC but with less than half of the computational reduction. BlockDrop for ResNet-32 prunes $54.0\%$ computation with $8.70\%$ error, $1.44\%$ higher than FGC for ResNet-32. Unlike layer pruning, FGC excavates redundancy of each gated layer in a fine-grained manner, maximally reducing irrelevant computation with minor representation degradation. FGC also consistently outperforms dynamic spatial pruning methods (SACT~\cite{DBLP:conf/eccv/XieZZHL20} and DynConv~\cite{Verelst2020DynamicCE}), which typically ignore the distribution consistency between instances and pruned architectures, by a large margin. FGC prunes similar computational costs with lower errors than SACT for ResNet-20 ($53.7\%$ reduction and $10.02\%$ error). Compared with DynConv for ResNet-32 ($51.3\%$ reduction and $7.47\%$ error), FGC prunes much more computation with slightly lower error.}

\textcolor{black}{
We further show that FGC achieves a superior accuracy-computation trade-off than BAS~\cite{ehteshami2020} and FBS~\cite{DBLP:conf/iclr/GaoZDMX19}, the very relevant dynamic channel pruning methods with gating modules. For ResNet-56, FGC prunes $3.3\%$ more computation than BAS with approximate classification error. For ResNet-20, FGC prunes $3.1\%$ more computation with lower error than BAS. FGC alleviates the inconsistency between distributions of features and gates in these methods. Our superior results validate the effectiveness of feature regularization toward gating modules.}

\textcolor{black}{When extending to WideResNet~\cite{WideResNet} and MobileNetV2~\cite{MobileNetV2}, FGC still achieves consistent improvements, as shown in Table~\ref{tab:more_networks}. On WRN-28-10, FGC prunes 3.6\% more computation while achieving 0.3\% less error. FGC could even further reduce redundancy in the lightweight network. On MobileNetV2, FGC prunes 4.5\% more computation and achieves 0.2\% less error. The results reflect FGC's generalizability on various network architectures.}

\begin{table}[ht]
    \scriptsize
    \centering
    \caption{Comparisons of the pruned WideResNet and MobileNetV2 on CIFAR10.}
    \begin{tabular}{cccc}
        \toprule
        Model & Method & Error (\%) $\downarrow$ & Pruning (\%)
        $\uparrow$ \\
        \midrule
        \multirow{3}{*}{WRN-28-10} & Baseline & 5.13 & 0 \\
         & w/o FGC & 6.15 & 42.5 \\
         & FGC & 5.85 & 46.1 \\
         \hline
        \multirow{3}{*}{MobileNetV2} & Baseline & 5.19 & 0 \\
         & w/o FGC & 6.27 & 38.8 \\
         & FGC & 6.07 & 43.3 \\
        \bottomrule
    \end{tabular}
    \label{tab:more_networks}
\end{table}

%%%%%%%%%%%%%%%%%%%%%%%%%%%%%%%%%%%%%%%%%%%%%%%
%%%%%%%%%%%%%%%   CIFAR100  %%%%%%%%%%%%%%%%%%%
%%%%%%%%%%%%%%%%%%%%%%%%%%%%%%%%%%%%%%%%%%%%%%%
\begin{table}[h]
%\small
\scriptsize
%\tiny
\centering
\setlength{\belowcaptionskip}{2pt}
\caption{Comparisons of the pruned ResNet on CIFAR100.}
\label{tab:cifar100}
\begin{tabular}{@{}ccccc@{}}
\toprule
Model    & Method     & Dynamic     & Error (\%) $\downarrow$    & Pruning (\%) $\uparrow$ \\ \midrule
\multirow{4}{*}{\rotatebox[origin=c]{90}{ResNet-20}} & Baseline  & -            & 31.38          & 0.0           \\
         & BAS~\cite{ehteshami2020}                 & \checkmark                & 31.98          & 26.2          \\
         & \textbf{FGC (Ours)}                      & \checkmark                & \textbf{31.37} & 26.9          \\
         & \textbf{FGC (Ours)}                      & \checkmark                & 31.69          & \textbf{31.9} \\ \midrule \midrule
\multirow{5}{*}{\rotatebox[origin=c]{90}{ResNet-32}} & Baseline  & -            & 29.85          & 0.0           \\
         & CAC~\cite{DBLP:journals/tnn/ChenXDLH21}  & \xmark                    & 30.49          & 30.1          \\
         & BAS~\cite{ehteshami2020}                 & \checkmark                & 30.03          & 39.7          \\
         & \textbf{FGC (Ours)}                      & \checkmark                & \textbf{29.90} & 40.4          \\
         & \textbf{FGC (Ours)}                      & \checkmark                & 30.09          & \textbf{45.1} \\ \midrule \midrule
\multirow{5}{*}{\rotatebox[origin=c]{90}{ResNet-56}} & Baseline  & -            & 28.44          & 0.0           \\
         & CAC~\cite{DBLP:journals/tnn/ChenXDLH21}  & \xmark                    & 28.69          & 30.0          \\
         & BAS~\cite{ehteshami2020}                 & \checkmark                & 28.37          & 37.1          \\
         & \textbf{FGC (Ours)}                      & \checkmark                & \textbf{27.87} & 38.1          \\
         & \textbf{FGC (Ours)}                      & \checkmark                & 28.00          & \textbf{40.7} \\ \bottomrule
\end{tabular}%
\end{table}
%%%%%%%%%%%%%%%%%%%%%%%%%%%%%%%%%%%%%%%%%%%%%%%

\textbf{CIFAR100.} In Table~\ref{tab:cifar100}, we present the pruning results of ResNet-\{20, 32, 56\}. For example, after pruning $26.9\%$ computational cost of ResNet-20, FGC even achieves $0.01\%$ less classification error. For ResNet-32 and ResNet-56, FGC outperforms SOTA methods CAC~\cite{DBLP:journals/tnn/ChenXDLH21} and BAS~\cite{ehteshami2020} with significant margins, from perspectives of both test error and computation reduction. Specifically, FGC, compared with BAS, reduces $5.4\%$ and $3.6\%$ more computation of ResNet-32 and ResNet-56 without bringing more errors. Experimental results on CIFAR100 justify the generality of our method to various datasets. 

%%%%%%%%%%%%%%%%%%%%%%%%%%%%%%%%%%%%%%%%%%%%%%%
%%%%%%%%%%%%%%%   ImageNet   %%%%%%%%%%%%%%%%%%
%%%%%%%%%%%%%%%%%%%%%%%%%%%%%%%%%%%%%%%%%%%%%%%
\begin{table}[t]
\scriptsize
\centering
\setlength{\belowcaptionskip}{2pt}
\caption{Comparisons of the pruned ResNet on ImageNet.}
\label{tab:imagenet}
\begin{tabular}{@{}ccccccc@{}}
\toprule
Model   & Method   & Dynamic & \textcolor{black}{Type} & Top-1 (\%) $\downarrow$ & Top-5 (\%) $\downarrow$ & Pruning (\%) $\uparrow$ \\ \midrule

\multirow{12}{*}{\rotatebox[origin=c]{90}{ResNet-18}} & Baseline & - & \textcolor{black}{-}  & 30.15  & 10.92 & 0.0  
\\
 & SFP~\cite{DBLP:conf/ijcai/HeKDFY18}        & \xmark  & \textcolor{black}{C}     & 32.90 & 12.22 & 41.8 \\
 & FPGM~\cite{DBLP:conf/cvpr/HeLWHY19}        & \xmark  & \textcolor{black}{C}     & 31.59 & 11.52 & 41.8 \\
 & PFP~\cite{DBLP:conf/iclr/LiebenweinBLFR20} & \xmark  & \textcolor{black}{C}     & 34.35 & 13.25 & 43.1 \\
 & DSA~\cite{DBLP:conf/eccv/NingZLLWY20}      & \xmark  & \textcolor{black}{CN}    & 31.39 & 11.65 & 40.0 \\
 & LCCN~\cite{DBLP:conf/cvpr/DongHYY17}       & \checkmark & \textcolor{black}{S}  & 33.67 & 13.06 & 34.6 \\
 & \textcolor{black}{AIG~\cite{DBLP:conf/eccv/VeitB18}}   & \textcolor{black}{\checkmark} & \textcolor{black}{L}  & \textcolor{black}{32.01} & \textcolor{black}{-}     & \textcolor{black}{22.3} \\ 
 & CGNet~\cite{DBLP:conf/nips/HuaZSZS19}      & \checkmark & \textcolor{black}{C}  & 31.70 & -     & \textbf{50.7} \\
 & FBS~\cite{DBLP:conf/iclr/GaoZDMX19}        & \checkmark & \textcolor{black}{C}  & 31.83 & 11.78 & 49.5 \\
 & BAS~\cite{ehteshami2020}                   & \checkmark & \textcolor{black}{C}  & 31.66 &  -    & 47.1  \\ \hline
 & \textbf{FGC (Ours)}   & \checkmark & \textcolor{black}{C} & \textbf{30.67} & \textbf{11.07}  & 42.2 \\
 & \textbf{FGC (Ours)}   & \checkmark & \textcolor{black}{C} & 31.49          & 11.62           & 48.1 \\  \bottomrule
\end{tabular}%
\end{table}
%%%%%%%%%%%%%%%%%%%%%%%%%%%%%%%%%%%%%%%%%%%%%%%

\textcolor{black}{
\textbf{ImageNet.} In Table~\ref{tab:imagenet}, we further compare FGC with methods on the large-scale dataset to prune ResNet-18. One can see that FGC outperforms most static pruning methods. FPGM~\cite{DBLP:conf/cvpr/HeLWHY19} reduces $41.8\%$ computation with $31.59\%$ top-1 error, while FGC reduces $42.2\%$ computation with $30.67\%$ top-1 error. It is comparable with other dynamic pruning methods. For example, compared with BAS~\cite{ehteshami2020}, FGC prunes more computation ($48.1\%$ $vs$ $47.1\%$) with lower top-1 error ($31.49\%$ $vs$ $31.66\%$). FGC outperforms AIG~\cite{DBLP:conf/eccv/VeitB18} and LCCN~\cite{DBLP:conf/cvpr/DongHYY17} significantly with much higher computation reduction and lower errors. Experimental results demonstrate our method's effectiveness in solving the complex representation redundancy on the large-scale image dataset.}

\subsection{\textcolor{black}{Transferring Tasks}}
\textcolor{black}{
The proposed gated models can be generalized to object detection and semantic segmentation tasks.}

%%%%%%%%%%%%%%%%%%%%%%%%%%%%%%%%%%
% Faster R-CNN, PASCAL VOC 2007
%%%%%%%%%%%%%%%%%%%%%%%%%%%%%%%%%%
\begin{table}[htp]
    \footnotesize
    \centering
    \setlength{\belowcaptionskip}{2pt}
    \caption{\textcolor{black}{Object detection on PASCAL VOC 2007 dataset with Faster R-CNN.}}
    \label{tab.detection}
        \textcolor{black}{
        \begin{tabular}{@{}cccc@{}}
        \toprule
        Backbone    & Method       & mAP (\%)     & Pruning (\%) \\ \midrule
        \multirow{3}{*}{\rotatebox[origin=c]{90}{ResNet-34}} & Baseline & 73.57 & 0 \\
                    & w/o FGC      & 67.66         & 50.6        \\
                    & FGC          & 69.18         & 52.0        \\ \midrule
        \multirow{3}{*}{\rotatebox[origin=c]{90}{ResNet-50}} & Baseline & 74.77 & 0 \\
                    & w/o FGC      & 71.38         & 34.3        \\
                    & FGC          & 71.81         & 34.3        \\
        \bottomrule
        \end{tabular}%
        }
\end{table}
%%%%%%%%%%%%%%%%%%%%%%%%%%%%%%%%%%

\textcolor{black}{
\textbf{Object Detection.} Experiments are conducted on PASCAL VOC 2007 using the Faster R-CNN method. The mAP (mean average precision) and the average pruning ratio are used to evaluate the accuracy-computation trade-off. The average pruning ratio is calculated based on the backbone. As shown in Table~\ref{tab.detection}, with a slight accuracy drop, FGC brings the detector with more than $50\%$ computation reduction. Note that FGC improves the trade-off compared with vanilla gated models. It achieves $0.48\%$ higher mAP with approximate computation of vanilla gated model on ResNet-50 backbone. And FGC achieves $1.52\%$ higher mAP with more reduction ($52.0\%$ vs $50.6\%$) on ResNet-34. These experiments indicate that our method could better utilize instance-wise redundancy to derive compact sub-networks for each input candidate.}

%%%%%%%%%%%%%%%%%%%%%%%%%%%%%%%%%%
% PSPNet, Cityscapes
%%%%%%%%%%%%%%%%%%%%%%%%%%%%%%%%%%
\begin{table}[htp]
    \footnotesize
    \centering
    \setlength{\belowcaptionskip}{2pt}
    \caption{\textcolor{black}{Semantic segmentation on Cityscapes dataset with PSPNet.}}
    \label{tab.segmentation}
        \textcolor{black}{
        \begin{tabular}{@{}ccccc@{}}
        \toprule
        Backbone    & Method    & mIoU (\%)     & PixelACC (\%)     & Pruning (\%)   \\ \midrule
        \multirow{4}{*}{\rotatebox[origin=c]{90}{ResNet-50}} & Baseline & 71.68 & 96.08 & 0    \\  
                    & BAS~\cite{ehteshami2020} & 71.90        & 93.50          & 23.7          \\     
                    & w/o FGC       & 71.70 (+0.02) & 96.18 (+0.10)   & 34.7        \\  
                    & FGC           & 72.32 (+0.64) & 96.18 (+0.10)   & 34.6        \\ \bottomrule
        \end{tabular}%
        }
\end{table}
%%%%%%%%%%%%%%%%%%%%%%%%%%%%%%%%%%

\textcolor{black}{
\textbf{Semantic Segmentation.} We conduct semantic segmentation experiments on Cityscapes using PSPNet. We use mean IoU (Intersection over Union), pixel accuracy, and average pruning ratio to evaluate gated models. As shown in Table~\ref{tab.segmentation}, the baseline model achieves $71.68\%$ IoU and $96.08\%$ pixel accuracy. Our gated model performs better segmentation results with significant computation reduction in comparison. Furthermore, FGC achieves $72.32\%$ IoU and $96.18\%$ pixel accuracy with $34.6\%$ computation reduction, outperforming the vanilla gated model, which achieves $71.70\%$ IoU with an approximate computational cost. It is concluded that our method achieves a better accuracy-computation trade-off.}

\section{Conclusions}
We propose a self-supervised feature-gate coupling (FGC) method to reduce the distortion of gated features by regularizing the consistency between feature and gate distributions. FGC takes the instance neighborhood relationships in the feature space as the objective while pursuing the instance distribution coupling with the gate space by aligning instance neighborhood relationships in the feature and gate spaces. FGC utilizes the $k$NN method to explore the instance neighborhood relationships in the feature space and utilizes CSL to regularize gating modules with the generated self-supervisory signals. Experiments validate that FGC improves DNP performance with striking contrast with the state-of-the-art methods. FGC provides fresh insight into the network pruning problem.
\textcolor{black}{When it comes to scenarios with long-tailed distribution data or incremental unseen classes, there may be difficulties in fitting the feature distributions for gate space alignment due to biased distribution estimates. Distribution calibration methods~\cite{DC2021, PDC2023} that leverage implicit class relationships may raise insights to alleviate the issue. We leave these interesting problems for future exploration.}
% \textcolor{black}{When it comes to scenarios with long-tailed distribution data or incremental unseen classes, there may be hard to fit the feature distributions for gate space alignment. We will leave these interesting problems for future exploration.}

\section*{Acknowledgment}
 This work was supported by National Natural Science Foundation of China (NSFC) under Grant 61836012, 61771447 and 62006216, the Strategic Priority Research Program of Chinese Academy of Sciences under Grant No. XDA27000000.

\bibliography{bibfile}

%%%%%%%%%%%%%%%%%%%%%%%%%%%%%%%%%%%%%

\end{document}